\documentclass[runningheads]{llncs}
\usepackage[utf8]{inputenc} 
\usepackage[T1]{fontenc}    
\usepackage{hyperref}       
\usepackage{url}            
\usepackage{booktabs}       
\usepackage{amsfonts}       
\usepackage{nicefrac}       
\usepackage{microtype}      
\usepackage[pdftex]{graphics,graphicx}
\usepackage{wrapfig}
\usepackage{amssymb}
\usepackage{amsmath}
\usepackage{xcolor}
\usepackage{microtype}
\usepackage{graphicx}
\usepackage{subfigure}
\usepackage{booktabs} 
\usepackage{dsfont}
\usepackage{float}

\newcommand{\RR}{\mathbb{R}}
\newcommand{\fNN}{f^{\text{\normalfont{NN}}}}
\newcommand{\subsuperNN}{{\text{\normalfont{NN}}}}
\newcommand{\fint}{f^{\text{\normalfont{int}}}}

\newcommand{\flin}{f^{\text{\normalfont{lin}}}}
\newcommand{\flinT}{f^{\text{\normalfont{lin}},T}}

\newcommand{\tst}{{\text{\normalfont test}}}
\newcommand{\QED}{\hfill\ensuremath{\square}}
\newcommand{\ii}{\mathds{1}}

\DeclareMathOperator*{\argmin}{arg\,min}
\usepackage{graphicx}
\begin{document}
\title{Which Minimizer Does My\\Neural Network Converge To?}
%

\author{Manuel Nonnenmacher\inst{1,2} \and
David Reeb\inst{1} \and
Ingo Steinwart\inst{2}}
\authorrunning{M. Nonnenmacher et al.}
%
\institute{Bosch Center for Artificial Intelligence (BCAI), 71272 Renningen, Germany, \email{manuel.nonnenmacher@de.bosch.com} \and
Institute for Stochastics and Applications, University of Stuttgart, 70569 Stuttgart, Germany}

%
\maketitle              
\begin{abstract}The loss surface of an overparameterized neural network (NN) possesses many global minima of zero training error. We explain how common variants of the standard NN training procedure change the minimizer obtained. First, we make explicit how the size of the initialization of a strongly overparameterized NN affects the minimizer and can deteriorate its final test performance. We propose a strategy to limit this effect. Then, we demonstrate that for adaptive optimization such as AdaGrad, the obtained minimizer generally differs from the gradient descent (GD) minimizer. This adaptive minimizer is changed further by stochastic mini-batch training, even though in the non-adaptive case, GD and stochastic GD result in essentially the same minimizer. Lastly, we explain that these effects remain relevant for less overparameterized NNs. While overparameterization has its benefits, our work highlights that it induces sources of error absent from underparameterized models.
\keywords{Overparameterization  \and Optimization \and Neural Networks.}
\end{abstract}
	\section{Introduction}

	Overparameterization is a key ingredient in the success of neural networks (NNs), thus modern NNs have become ever more strongly overparameterized. As much as this has helped increase NN performance, overparameterization has also caused several puzzles in our theoretical understanding of NNs, especially with regards to their good optimization behavior \cite{zhang2019identity} and favorable generalization properties \cite{zhang2016understanding}. In this work we shed light on the optimization behavior, identifying several caveats.

	More precisely, we investigate how the obtained minimizer can change depending on the NN training procedure -- we consider common techniques like adjusting the initialization size, the use of adaptive optimization, and stochastic gradient descent (SGD).

\begin{figure}[ht]
		\centering
		\begin{minipage}[b]{0.7\textwidth}
		\includegraphics[width=\textwidth]{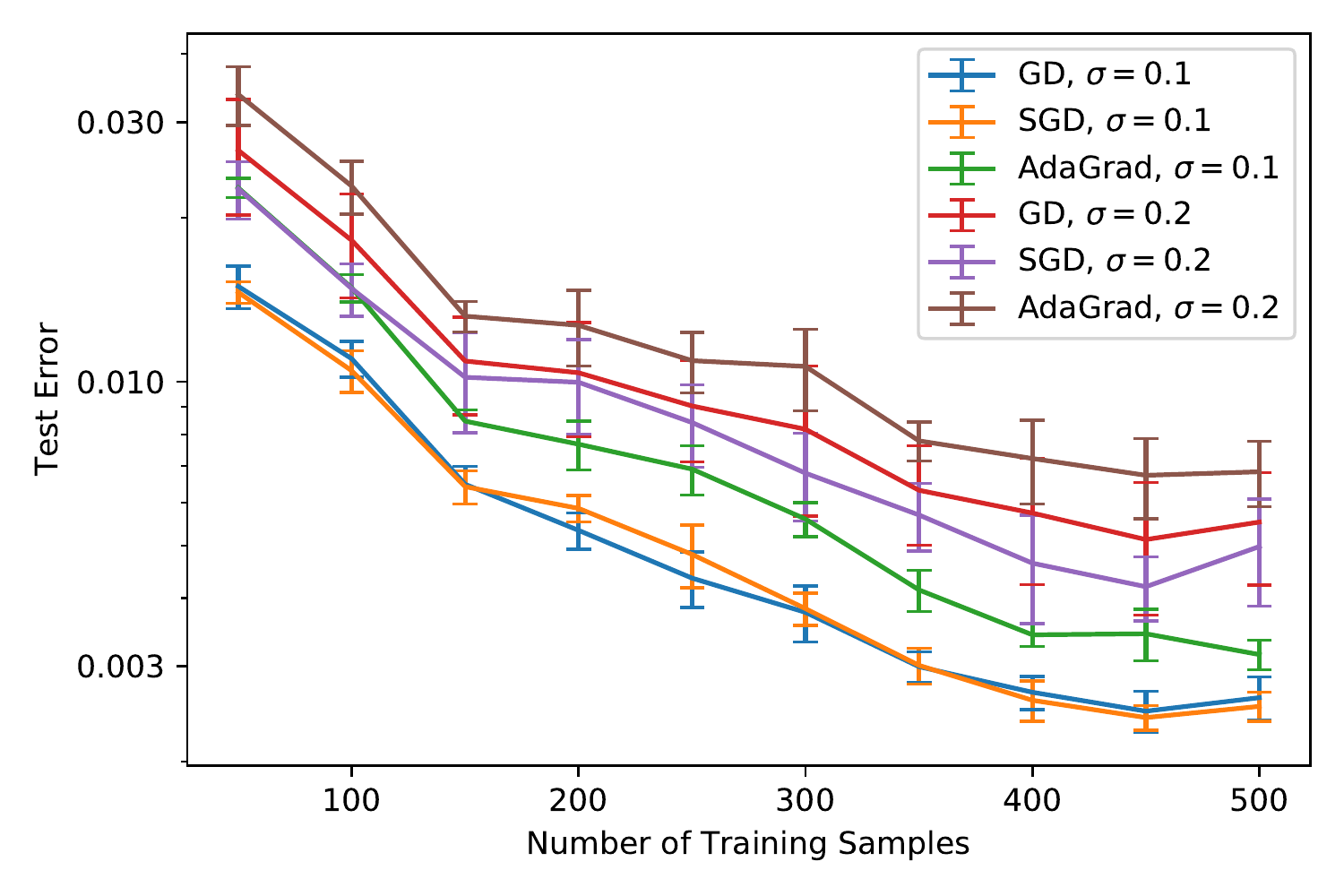}
		\end{minipage}
		\caption{The test performance of an overparameterized NN depends considerably on the optimization method (GD, SGD, AdaGrad) and on the initialization size $\sigma$, even though all nets have been trained to the same low empirical error of $10^{-5}$. Shown are results on MNIST 0 vs.\ 1 under squared loss in 5 repetitions of each setting, varying the degree of overparameterization by changing the training set size. Our theoretical results explain and quantify these differences between training choices.
		}
		\label{fig:intro}
	\end{figure}

These training choices can have significant impact on the final test performance, see Fig.\ \ref{fig:intro}. While some of these peculiar effects had been observed experimentally \cite{wilson2017marginal,zhang2019type},

we explain and quantify them in a general setting.
Note further that this effect is absent from the more commonly studied underparameterized models, whose minimizer is generically unique (\cite{shah2018minimum} and App.\ \ref{app:underpara_model}).

Our analysis makes use of the improved understanding of the training behavior of strongly overparameterized NNs which has been achieved via the Neural Tangent Kernel (NTK) \cite{jacot2018neural,lee2019wide,du2018gradient}.
Through this connection one can show that overparameterized NNs trained with gradient descent (GD) converge to a minimizer which is an interpolator of low complexity w.r.t.\ the NTK \cite{belkin2018understand}. We also extend our analysis to less overparameterized NNs by using linearizations at later training times instead of the NTK limit (Sect.\ \ref{not-stronly-overparam-sect}).

Our contributions are as follows:
	\begin{itemize}
		\item We explain quantitatively how the size of initialization impacts the trained overparameterized NN and its test performance. While the influence can generally be severe, we suggest a simple algorithm to detect and mitigate this effect (Sect.\ \ref{sect:init}).
		\item We prove that the choice of adaptive optimization method changes the minimizer obtained and not only the training trajectory (Sect.\ \ref{sect:adaptive}). This can significantly affect the test performance. As a technical ingredient of independent interest we prove that strongly overparameterized NNs admit a linearization under adaptive training similar to GD and SGD training \cite{lee2019wide,dudeep2018gradient,allen2018convergence}.
		\item We show that the batch size of mini-batch SGD affects the minimizer of adaptively trained NNs, 
in contast to the non-adaptive setting, where the SGD minimizer is virtually the same as in full-batch (GD) training (Sect.\ \ref{sect:sgd}).
		\item Our theoretical findings are confirmed by extensive experiments on different datasets, where we investigate the effect of the changed minimizer on the test performance (Sect.\ \ref{sect:experiments}).
	\end{itemize}

\section{Background}
\label{sec:notation}
	Throughout, the $N$ training points $\mathcal{D}=\{(x_i,y_i)\}_{i=1}^N\subset B_1^d(0)\times\RR$ have inputs from the $d$-dimensional unit ball $B_1^d(0):=\{ x\in\RR^d:\|x\|_2 \leq 1\}$, and one-dimensional outputs for simplicity. $\mathcal{D}$ is called \emph{non-degenerate} if $x_i \nparallel x_j$ for $i\neq j$.
	We often view the training inputs $X=(x_1,\ldots,x_N)^T\in\RR^{N\times d}$ and corresponding labels $Y=(y_1,\ldots,y_N)^T\in\RR^N$ in matrix form.
	For any function $g$ on $\RR^d$, we define $g(X)$ by row-wise application of $g$.

	The output of a fully-connected NN with $L$ layers and parameters $\theta$ is denoted $\fNN_\theta(x) = h^L(x)$ with
	\begin{equation}
	\begin{aligned}\label{NNparametrization}
	h^l(x) &~=~\frac{\sigma}{\sqrt{m_l}}\, W^l a(h^{l-1})+\sigma b^l \quad\text{for}~l=2,\ldots,L, \\ h^1(x) &~=~ \frac{\sigma}{\sqrt{m_1}}\, W^1x + \sigma b^1,
	\end{aligned}
	\end{equation}
	where $a: \mathbb{R} \to \mathbb{R}$ is the activation function, applied component-wise. We assume $a$ either to have bounded second derivative  \cite{dudeep2018gradient} or to be the ReLU function $a(h)=\max\{h,0\}$ \cite{allen2018convergence}. Layer $l$ has $m_l$ neurons ($m_0=d$, $m_L=1$); we assume $m_l\equiv m$ constant for all hidden layers $l=1,\ldots,L-1$ for simplicity.\footnote{For unequal hidden layers, the infinite-width limit (below) is $\min_{1\leq l\leq L-1}\{m_l\}\to\infty$ \cite{lee2019wide}.} $W^l \in \mathbb{R}^{m_l \times m_{l-1}}$ and $b^l \in \mathbb{R}^{m_l}$ are the NN weights and biases, for a total of $P = \sum_{l=1}^{L}(m_{l-1}+1)m_l$ real parameters in $\theta = [W^{1:L},b^{1:L}]$. We keep the parameter scaling $\sigma$ as an explicit scalar parameter, that can be varied \cite{jacot2018neural}. The parametrization (\ref{NNparametrization}) together with a \emph{standard normal initialization} $W^l_{i,j},b^l_{i} \sim \mathcal{N}(0,1)$ (sometimes with \emph{zero biases} $b_i^l=0$) is the \emph{NTK-parametrization} \cite{dudeep2018gradient}.
	This is equivalent to the standard parametrization (i.e., no prefactors in (\ref{NNparametrization})) and initialization $W^l_{i,j}\sim\mathcal{N}(0,\sigma^2/m_{l})$, $b^l_{i}\sim\mathcal{N}(0,\sigma^2)$ \cite{he2015delving} in a NN forward pass,
	while in gradient training the two parametrizations differ by a width-dependent scaling factor of the learning rate \cite{jacot2018neural,lee2019wide}.
	
	We mostly consider the squared error $|\hat{y}-y|^2/2$ and train by minimizing its empirical loss
	\begin{equation*}
	L_{\mathcal D}(f_\theta) =\frac{1}{2N}\sum_{(x,y) \in \mathcal{D}} |f_{\theta}(x)-y|^2  =\frac{1}{2N}\|f_{\theta}(X)-Y\|_2^2.
	\end{equation*}
	We train by \emph{discrete} update steps, i.e.\ starting from initialization $\theta=\theta_0$ the parameters are updated via the discrete iteration $\theta_{t+1} = \theta_t - \eta U_t[\theta_t]$ for $t=0,1,\ldots$, where $U_t$ is some function of the (past and present) parameters and $\eta$ the learning rate. Gradient descent (GD) training uses the present loss gradient $U_t[\theta_t] = \left.\nabla_{\theta}L_{\mathcal D}(f_{\theta})\right|_{\theta=\theta_t}$; for adaptive and stochastic gradient methods, see Sects.\ \ref{sect:adaptive} and \ref{sect:sgd}.
	
	A central object in our study is the feature map $\phi(x):= \nabla_{\theta}\fNN_{\theta}(x)\big|_{\theta=\theta_0} \in \mathbb{R}^{1\times M}$ associated with a NN $\fNN_\theta$ at its initialization $\theta_0$. With this we will consider the NN's linearization around $\theta_0$ as
	\begin{equation}\label{flin-definition}
	\flin_\theta(x)~:=~\fNN_{\theta_0}(x)+\phi(x)(\theta-\theta_0).
	\end{equation}
	On the other hand, $\phi$ gives rise to the so-called neural tangent kernel (NTK) $K:\RR^d\times\RR^d\to\RR$ by $K(x,x'):=\phi(x)\phi(x')^T$ \cite{jacot2018neural,lee2019wide}. We use the associated kernel norm to define the \emph{minimum complexity interpolator} of the data $\mathcal{D}$:
	\begin{equation}\label{definition-fint}
	\fint := \argmin_{f \in \mathcal H_K} \left\lVert f \right\rVert_{\mathcal{H}_K}\quad\text{ subject to~~}Y= f(X),
	\end{equation}
	where $\mathcal{H}_K$ is the reproducing kernel Hilbert space (RKHS) associated with $K$ \cite{belkin2018understand}. Its explicit solution is $\fint(x)=\phi(x)\phi(X)^T\left(\phi(X)\phi(X)^T\right)^{-1}Y$ (App.\ \ref{app:interpolating_kernel}). Here, $\phi(X)\phi(X)^T=K(X,X)$ is invertible for ``generic'' $X$ and $\theta_0$ in the overparameterized regime $P\geq N$. Technically, we always assume that the infinite-width kernel $\Theta(x,x'):=\lim_{m\to\infty}K(x,x')$ has positive minimal eigenvalue $\lambda_0:=\lambda_{\min}(\Theta(X,X))>0$ on the data (this $\lim_{m\to\infty}$ exists in probability \cite{jacot2018neural}). $\lambda_0>0$ holds for non-degenerate $\mathcal{D}$ and standard normal initialization with zero biases \cite{dudeep2018gradient}; for standard normal initialization (with normal biases) it suffices that $x_i\neq x_j$ for $i\neq j$.
	
	With these prerequisites, we use as a technical tool the fact that a strongly overparameterized NN stays close to its linearization during GD training (for extensions, see Thm.\ \ref{thm:adaptivelinear}). 
	More precisely, the following holds:
	\begin{lemma}\label{lemma:lee2019wide} \textbf{\emph{(\cite{lee2019wide,allen2018convergence})}} Denote by $\theta_t$ and $\tilde{\theta}_t$ the parameter sequences obtained by gradient descent on the NN $\fNN_\theta$ (\ref{NNparametrization}) and on its linearization $\flin_{\tilde{\theta}}$ (\ref{flin-definition}), respectively, starting from the same initialization $\theta_0=\tilde{\theta}_0$ and with sufficiently small step size $\eta$. There exists some $C={\mathrm{poly}}(1/\delta,N,1/\lambda_0,1/\sigma)$ such that for all $m \geq C$ and for all $x\in B_1^d(0)$ it holds with probability at least $1-\delta$ over the random initialization $\theta_0$ that: $\sup_{t} |\fNN_{\theta_t}(x) - \flin_{\tilde{\theta}_t}(x)|^2 \leq O(1/m)$.
	\end{lemma}

	\section{Impact of initialization}\label{sect:init}
	In this section we quantify theoretically how the initialization $\theta_0$ influences the final NN trained by gradient descent (GD), and in particular its test error or risk. As a preliminary result, we give an analytical expression for the GD-trained NN, which becomes exact in the infinite-width limit:
	\begin{theorem}\label{thm:fNNfint}
		Let $\fNN$ be the fully converged solution of an $L$-layer ReLU-NN (\ref{NNparametrization}), trained by gradient descent under squared loss on non-degenerate data ${\mathcal D}=(X,Y)$. There exists $C = {\mathrm{poly}}(1/\delta,N,1/\lambda_0,1/\sigma)$ such that whenever there are $m \geq C$ neurons in each hidden layer, it holds for any $x \in B_1^d(0)$ that, with probability at least $1-\delta$ over standard normal initialization $\theta_0$,
		\begin{align}\label{theo:keyformula}
			\fNN(x) ~=&~ \phi(x) \phi(X)^T\big(\phi(X)\phi(X)^T\big)^{-1}Y + \frac{1}{L}\phi(x)\big(\ii-P_{\mathcal{F}}\big)\theta_0 \,+\, O\Big(\frac{1}{\sqrt{m}}\Big),
		\end{align}
		where $P_{\mathcal{F}}:=\phi(X)^T\big(\phi(X)\phi(X)^T\big)^{-1} \phi(X)$ is the projector onto the data feature subspace.
	\end{theorem}
	Results similar to Thm.\ \ref{thm:fNNfint} have been found in other works before \cite{jacot2018neural,lee2017deep,zhang2019type}. Our result has a somewhat different form compared to them and makes the dependence on the initialization $\theta_0$ more explicit. For this, we simplified the $Y$-independent term by using the property $\fNN_\theta(x)=\frac{1}{L}\langle \theta , \nabla_\theta \fNN_\theta(x)\rangle$ of ReLU-NNs (Lemma \ref{lemma:reluexpression}). Further, our Thm.\ \ref{thm:fNNfint} is proven for discrete, rather than continuous, update steps. The proof in App.\ \ref{app:proof_fNNfint} first solves the dynamics of the linearized model (\ref{flin-definition}) iteratively and recovers the first two terms in the infinite training time limit. Finally, Lemma \ref{lemma:lee2019wide} gives $O(1/\sqrt{m})$-closeness to $\fNN(x)$ in the strongly overparameterized regime.
	
	The expression (\ref{theo:keyformula}) for the converged NN has two main parts. The first term is just the minimum complexity interpolator $\fint(x)=\phi(x) \phi(X)^T(\phi(X)\phi(X)^T)^{-1}Y$ from Eq.\ (\ref{definition-fint}), making the solution interpolate the training set $\mathcal{D}$ perfectly. Note, $\fint(x)$ is virtually independent of the random initialization $\theta_0$ in the strongly overparameterized regime since $\phi(x)\phi(x')^T$ converges in probability as $m\to\infty$ (Sect.\ \ref{sec:notation}); intuitively, random $\theta_0$'s yield similarly expressive features $\phi(x)\in\mathbb{R}^P$ as $P\to\infty$.
	
	The second term in (\ref{theo:keyformula}), however, depends on $\theta_0$ explicitly. More precisely, it is proportional to the part $(\ii\!-\!P_{\mathcal{F}})\theta_0$ of the initialization that is orthogonal to the feature subspace $\mathcal{F}={\mathrm{span}}\{\phi(x_1),\ldots,\phi(x_N)\}$ onto which $P_{\mathcal{F}}$ projects. This term is present as GD alters $\theta_t\in\RR^P$ only along the $N$-dimensional $\mathcal{F}$. It vanishes on the training inputs $x=X$ due to $\phi(X)(\ii-P_{\mathcal{F}})=0$.
	
	Our main concern is now the extent to which the \emph{test} error is affected by $\theta_0$ and thus in particular by the second term $\phi(x)(\ii\!-\!P_{\mathcal{F}})\theta_0/L$. Due to its $Y$-independence, this term will generally harm test performance. While this holds for large initialization scaling $\sigma$, small $\sigma$ suppresses this effect:
	\begin{theorem}\label{thm:initsigma}
		Under the prerequisites of Thm.\ \ref{thm:fNNfint} and fixing a test set $\mathcal{T}=(X_{\mathcal{T}},Y_{\mathcal{T}})$ of size $N_{\mathcal{T}}$, there exists $C = {\mathrm{poly}}(N,1/\delta,1/\lambda_0,1/\sigma,N_{\mathcal{T}})$ such that for $m\geq C$ the test error $L_{\mathcal{T}}(\fNN)$ of the trained NN satisfies the following bounds, with probability at least $1- \delta$ over the standard normal initialization with zero biases,
		\begin{equation}\label{eq:thmtestbounds}
		\begin{split}
		&\sqrt{L_{\mathcal{T}}({\fNN})}\geq\sigma^{L} J(X_{\mathcal{T}}) - \sqrt{L_{\mathcal{T}}(\fint)} -  O\Big(\frac{1}{\sqrt{m}}\Big)\,, \\ 
&\sqrt{L_{\mathcal{T}}({\fNN})} \leq\sqrt{L_{\mathcal{T}}(\fint)} + O\Big(\sigma^L+\frac{1}{\sqrt{m}}\Big),
        \end{split}
		\end{equation}
		where $J(X_{\mathcal{T}})$ is independent of the initialization scaling $\sigma$ and $J(X_{\mathcal{T}})>0$ holds almost surely.
		With standard normally initialized biases, the same bounds hold with both $\sigma^L$ replaced by $\sigma$.
	\end{theorem}
	The lower bound in (\ref{eq:thmtestbounds}) shows that big initialization scalings $\sigma$ leave a significant mark $\sim\sigma^L$ on the the test error, while the final \emph{training} error of $\fNN$ is always $0$ due to strong overparameterization. This underlines the importance of good initialization schemes, as they do not merely provide a favorable starting point for training, but impact which minimizer the weights converge to.
	To understand the scaling, note that the features scale with $\sigma$ like $\phi_\sigma(x)=\sigma^L\phi_1(x)$ (the behavior is more complex for standard normal biases, see App.\ \ref{app:proof_initbelow}. The first term in (\ref{theo:keyformula}) is thus invariant under $\sigma$, while the second scales as $\sim\sigma^L$.
	
	The main virtue of the upper bound in (\ref{eq:thmtestbounds}) is that the harmful influence of initialization can be reduced by adjusting $\sigma$ and $m$ simultaneously. To show this, App.\ \ref{app:proof_boundinit} takes care to bound the second term in (\ref{theo:keyformula}) on the test set $\|\phi(X_{\mathcal{T}})(1-P_{\mathcal{F}})\theta_0\|/\sqrt{N_\mathcal{T}}\leq O(\sigma^L)$ \emph{independently} of $\phi$'s dimension $P$, which would grow with $m$. Note further that the kernel interpolator $\fint$ in (\ref{eq:thmtestbounds}) with loss $L_{\mathcal{T}}(\fint)$ was recently found to be quite good empirically \cite{arora2019exact} and theoretically \cite{liang2018just,arora2019fine}.
	
	Based on these insights into the decomposition (\ref{theo:keyformula}) and the scaling of $L_{\mathcal{T}}(\fNN)$ with $\sigma$, 
we suggest the following algorithm to mitigate the potentially severe influence of the initialization on the test error in large NNs as much as possible: \emph{(a)} randomly sample an initialization $\theta_0$ and train ${\fNN}'$ using a standard scaling $\sigma'\simeq O(1)$ (e.g.\ \cite{he2015delving}); \emph{(b)} train ${\fNN}''$ with the same $\theta_0$ and a somewhat smaller $\sigma''<\sigma'$ (e.g.\ by several ten percent); \emph{(c)} compare the losses on a validation set ${\mathcal V}$: if $L_{\mathcal{V}}({\fNN}'')\approx L_{\mathcal{V}}({\fNN}')$ then finish with step (e), else if $L_{\mathcal{V}}$ decreases by a significant margin then continue; \emph{(d)} repeat from step (b) with successively smaller $\sigma'''<\sigma''$ until training becomes impractically slow; \emph{(e)} finally, return the trained $\fNN$ with smallest validation loss.
	
	It is generally not advisable to start training (or the above procedure) with a too small $\sigma<O(1)$ due to the vanishing gradient problem which leads to slow training, even though the upper bound in (\ref{eq:thmtestbounds}) may suggest very small $\sigma$ values to be beneficial from the viewpoint of the test error. A ``antisymmetrical initialization'' method was introduced in \cite{zhang2019type} to reduce the impact of the initialization-dependence on the test performance by doubling the network size; this however also increases the computational cost.
	
	Note further that the above theorems do not hold exactly anymore for $\sigma$ too small due to the $m\geq{\mathrm{poly}}(1/\sigma)$ requirement; our experiments (Fig.\ \ref{fig:intro} and Sect.\ \ref{sect:experiments}) however confirm the predicted $\sigma$-scaling even for less strongly overparameterized NNs.

	\section{Impact of adaptive optimization}\label{sect:adaptive}
	We now explain how the choice of adaptive optimization method affects the minimizer to which overparameterized models converge. The discrete weight update step for adaptive gradient training methods is
	\begin{equation}\label{adaptive_update}
	\theta_{t+1} = \theta_t - \eta D_t \nabla_\theta L_{\mathcal{D}}(\theta)\big|_{\theta=\theta_t},
	\end{equation}
	where the ``adaptive matrices'' $D_t \in \RR^{P\times P}$ are prescribed by the method. This generalizes GD, which is obtained by $D_t\equiv\ii$, and includes AdaGrad \cite{duchi2011adaptive} via $D_t = \Big(\text{diag}\big(\sum_{u=0}^{t}g_ug_u^T\big)\Big)^{-1/2}$ with the loss gradients $g_t = \nabla_{\theta} L(\theta)\big|_{\theta_t} \in \RR^P$, as well as {RMS\-prop} \cite{RmspropHinton}
	and other adaptive methods. 
	We say that a sequence of adaptive matrices concentrates around $D\in\RR^{P\times P}$ if there exists $Z \in \mathbb{R}$ such that $\| D_t-D\|_{\text{op}}/ D_{\max} \leq Z/\sqrt{m}$ holds for all $t\in\mathbb{N}$, where $D_{\max} := \sup_t \|D_t\|_{\text{op}}$; this is the simplest assumption under which we can generalize Thm.\ \ref{thm:fNNfint}, but in general we only need that the linearization during training holds approximately (App.\ \ref{app:proof_generaladaptive}).
	
	The following result gives a closed-form approximation to strongly overparameterized NNs during training by Eq.\ (\ref{adaptive_update}). Note that this overparameterized adaptive case was left unsolved in \cite{shah2018minimum}.
The closed-form expression allows us to illustrate via explicit examples that the obtained minimizer can be markedly different from the GD minimizer, see Example \ref{ex:1} below.
	\begin{theorem}\label{thm:generaladaptive}
		Given a NN $\fNN_{\theta}$ (\ref{NNparametrization}) and a non-degenerate training set ${\mathcal D}=(X,Y)$ for adaptive gradient training (\ref{adaptive_update}) under squared loss with adaptive matrices $D_t \in{\mathbb{R}}^{P\times P}$ concentrated around some $D$, there exists $C= {\mathrm{poly}}(N,1/\delta,1/\lambda_0,1/\sigma)$ such that for any width $m \geq C$ of the NN and any $x \in B_1^d(0)$ it holds with probability at least $1-\delta$ over the random initialization that
\begin{equation}\label{general_adaptive_formula}
\begin{split}
\fNN_{\theta_t}(x)~=&~\phi(x)A_t\left[\ii\ -\prod_{u=t-1}^{0}\big(\ii-\frac{\eta}{N}\phi(X) D_u\phi(X)^T\big)\right]\left(Y-\fNN_{\theta_0}(X)\right), \\
&+\fNN_{\theta_0}(x)+\phi(x)B_t+O(\frac{1}{\sqrt{m}}),
\end{split}
\end{equation}
where $A_t = D_{t-1}\phi(X)^T\left(\phi(X) D_{t-1} \phi(X)^T\right)^{-1}$ and
\begin{align}\nonumber
B_t = \sum_{v=2}^{t}(A_{v-1}\!-\!A_v)\!&\cdot\!\left[\ii\!\! -\!\!\!\!\prod_{w=v-2}^{0}\!\!\!\!\big(\ii\!\!-\!\!\frac{\eta}{N} \phi(X)D_w\phi(X)^T\big) \right]\cdot\left( Y-\fNN_{\theta_0}(X)\right)\!.\nonumber
\end{align}
	\end{theorem}
	To interpret this result, notice that on the training inputs $X$ we have $\phi(X)A_t=\ii$ and therefore $\phi(X)B_t=0$, so that the dynamics $\fNN_{\theta_t}(X)$ on the training data simplifies significantly: When the method converges, i.e.\ $\prod_{u=t-1}^0(\ldots)\to0$ as $t\to\infty$, we have $\fNN_{\theta_t}(X)\to Y+O(1/\sqrt{m})$, meaning that the training labels are (almost) perfectly interpolated at convergence. Even at convergence, however, the interpolating part $\fNN_{\theta_0}(x)+\phi(x)A_t(Y-\fNN_{\theta_0}(X))$ depends on $D_t$ (via $A_t$) for test points $x$. In addition to that, the term $\phi(x)B_t$ in (\ref{general_adaptive_formula}) is ``path-dependent'' due to the sum $\sum_{v=2}^t(\ldots)$ and takes account of changes in the $A_t$ (and thus, $D_t$) matrices during optimization, signifying changes in the geometry of the loss surface. The proof of Thm.\ \ref{thm:generaladaptive} in App.\ \ref{app:proof_generaladaptive} is based on a generalization of the linearization Lemma \ref{lemma:lee2019wide} for strongly overparameterized NNs to adaptive training methods (Thm.\ \ref{thm:adaptivelinear} in App.\ \ref{app:linearizationadaptive}).

	We make the dependence of the trained NN (\ref{general_adaptive_formula}) on the choice of adaptive method explicit by the following example of AdaGrad training. 
	\begin{example}\label{ex:1}
		Perform adaptive optimization on a ReLU-NN with adaptive matrices $D_t$ according to the AdaGrad prescription with 
		\begin{equation}\label{ada:assumption}
		g_t = a_t g \quad \text{where} \quad g \in \mathbb{R}^P,~a_t \in \mathbb{R},
		\end{equation}
		i.e.\ we assume that the gradients point all in the same direction $g$ with some decay behavior set by the $a_t$. We choose this simple setting for illustration purposes, but it occurs e.g.\ for heavily overparameterized NNs with $N=1$ training point, where $g=\phi(x_1)^T$ (a similar setting was used in \cite{zhang2019identity}). The adaptive matrices are then $D_t=D\left(\sum_{i=0}^t a_i^2\right)^{-1/2}$ with $D=\left({\mathrm{diag}}(gg^T)\right)^{-1/2}$.  As $D_t$ evolves only with scalar factors, $A_t=D\phi(X)^T\left(\phi(X)D\phi(X)^T\right)^{-1}$ is constant and thus $B_t=0$. We can then explicitly evaluate Thm.\ \ref{thm:generaladaptive} and use similar arguments as in Thm.\ \ref{thm:initsigma} to collect the $\fNN_{\theta_0}$-terms into $O(\sigma^L)$, to write down the minimizer at convergence:
		\begin{align} \label{adagrad_solution}
		\fNN(x)  ~=~&  \phi(x)D\phi(X)^T\left(\phi(X) D \phi(X)^T\right)^{-1}Y +  O\left(\sigma^L\right)+O\left(1/\sqrt{m}\right). 
		\end{align}
	\end{example}
	This example shows explicitly that the minimizer obtained by adaptive gradient methods in overparameterized NNs can be different from the GD minimizer, which results by setting $D=\ii$. This difference is not seen on the training inputs $X$, where Eq.\ (\ref{adagrad_solution}) always evaluates to $Y$, but only at test points $x$. In fact, any function of the form $\fNN(x)=\phi(x)w+O(\sigma^L+1/\sqrt{m})$ that interpolates the data, $\phi(X)w=Y$, can be obtained as the minimizer by a judicious choice of $D$. In contrast to the toy example in Wilson et al. \cite{wilson2017marginal}, our Example \ref{ex:1} does not require a finely chosen training set and is just a special case of the more general Thm.\ \ref{thm:generaladaptive}.

	Another way to interpret Example \ref{ex:1} is that this adaptive method converges to the interpolating solution of minimal complexity w.r.t.\ a kernel $K_D(x,x')=\phi(x)D\phi(x')^T$ different from the kernel $K(x,x')=\phi(x)\phi(x')^T$ associated with GD (see Sect.\ \ref{sec:notation} and Thm.\ \ref{thm:fNNfint}). Thus, unlike in Sect.\ \ref{sect:init} where the disturbing term can in principle be diminished by initializing with small variance, adaptive training directly changes the way in which we interpolate the data and is harder to control.

	Note that the situation is different for underparameterized models, where in fact the same (unique) minimizer is obtained irrespectively 
	of the adaptive method chosen (see App.\ \ref{app:underpara_model} and \cite{shah2018minimum}).

	\section{Impact of stochastic optimization}\label{sect:sgd}
	Here we investigate the effect SGD has on the minimizer to which the NN converges. The general update step of adaptive mini-batch SGD with adaptive matrices $D_t\in \RR^{P\times P}$ is given by
	\begin{equation}\label{generalsgdstep}
	\theta_{t+1} = \theta_t-\eta D_t \nabla_{\theta}L_{\mathcal{D}_{B_t}}(\theta)\big|_{\theta= \theta_t},
	\end{equation}
	where $\mathcal{D}_{B_t}$ contains the data points of the $t$-th batch $B_t$. Further, we denote by $X_{B_t}$ the corresponding data matrix obtained by zeroing all rows outside $B_t$. Ordinary SGD corresponds to $D_t= \ii$.
	
	The main idea behind the results of this section is to utilize the fact that ordinary SGD can be written in the form of an adaptive update step by using the adaptive matrix $D_{B_t}:= \frac{N}{|B_t|}\left(\phi(X_{B_t})\right)^T\left(\phi(X)\phi(X)^T\right)^{-1}\phi(X)$ in Eq.\ (\ref{adaptive_update}). Combining this re-writing of mini-batch SGD with the adaptive update rule, we can write Eq.\ (\ref{generalsgdstep}) as $\theta_{t+1} = \theta_t-\eta D_t D_{B_t} \nabla_{\theta}L_{\mathcal{D}}(\theta)|_{\theta= \theta_t}$. Now applying a similar approach as for Thm.\ \ref{thm:generaladaptive} leads to the following result:
	
	\begin{theorem}[informal]\label{thm:sgdadaptiveinformal}Let $\fNN_{adGD}$ and $\fNN_{adSGD}$ be fully trained strongly overparameterized NNs trained on the empirical squared loss with adaptive GD and adaptive SGD, respectively. Then, for sufficiently small learning rate $\eta$ it holds with high probability that:
		\begin{itemize}
			\item[(a)]If $D_t={\mathrm{const}}$, then $\big|\fNN_{adGD}(x)-\fNN_{adSGD}(x)\big|\leq O(1/\sqrt{m})$.
			\item[(b)]If $D_t$ changes during training, the minimizers $\fNN_{adGD}$ and $\fNN_{adSGD}$ differ by a path- and batch-size-dependent contribution on top of the $O(1/\sqrt{m})$ linearization error.
		\end{itemize}
	\end{theorem}
	
	Part (a) shows that GD and SGD lead to basically the same minimizer if the adaptive matrices $D_t$ do not change during training. This is the case in particular for vanilla (S)GD, where $D_t = \ii$. Part (b) on the other hand shows that for adaptive methods with varying adaptive matrices, the two NN minimizers obtained by GD and mini-batch SGD differ by a path-dependent contribution, where the path itself can be influenced by the batch size. We expect this effect to be smaller for more overparameterized models since then the adaptive matrices are expected to be more concentrated. For the formal version of Thm.\ \ref{thm:sgdadaptiveinformal} see App.\ \ref{app:SGDdetails}.

	One of the prerequisites of Thm.\ \ref{thm:sgdadaptiveinformal} is a small learning rate, but it is straightforward to generalize the results to any strongly overparameterized NN (in the NTK-regime) with a learning rate schedule such that the model converges to a minimizer of zero training loss (see App.\ \ref{app:proof_generalsgd}). 
	
	\section{Beyond strong overparameterization}
\label{not-stronly-overparam-sect}
	The previous three sections explain the impact of common training techniques on the obtained minimizer for strongly overparameterized NNs. In practice, NNs are usually less overparameterized or trained with a large initial learning rate, both of which can cause $\phi(X)$, and thus also $K(X,X)$, to change appreciably during training. This happens especially during the initial stages of training, where weight changes are most significant. The question thus arises how the theoretical results of Sects.\ \ref{sect:init}--\ref{sect:sgd} transfer to less overparameterized NNs. (Note that experimentally, the effects do still appear in less overparameterized NNs, see Fig.\ \ref{fig:intro}.)

The basis of our theoretical approach is the validity of Lemma \ref{lemma:lee2019wide} and its cousins like Thm.\ \ref{thm:adaptivelinear}, which build on the fact that the weights of a strongly overparameterized NN do not change significantly during training, i.e.\ $\|\theta_t-\theta_0\|_2$ remains small for all $t>0$. For less overparameterized NNs this does not hold in general. One can circumvent this by selecting a later training iteration $T>0$ such that $\|\theta_t-\theta_T\|_2$ is small enough for all $t>T$. One can always find such $T$, assuming that $\theta_t$ converges as $t\to\infty$. Next, to proceed with a similar analysis as before, we linearize the NN around iteration $T$ instead of Eq.\ (\ref{flin-definition}):
	\begin{equation}\label{flin-definition-later-iteration}
	\flinT_{\theta}(x)~:=~\fNN_{\theta_T}(x)+\phi_T(x)(\theta-\theta_T),
	\end{equation}
where $\phi_T(x):=\nabla_{\theta}\fNN_{\theta}(x)\big|_{\theta=\theta_T}$ are the NN features at training time $T$, assumed such that $\|\theta_t-\theta_T\|_2$ is sufficiently small for all $t>T$. Assuming further that $\lambda^T_0 := \lambda_{\min}(\phi_T(X)\phi_T(X)^T)>0$, gives straightforward adaptations of Thms.\ \ref{thm:fNNfint}, \ref{thm:generaladaptive}, and \ref{thm:sgdadaptiveinformal} with features $\phi_T(x)$ and valid at times $t\geq T$. The main difference is that the results are no longer probabilistic but rather conditional statements.

To demonstrate how this observation can be applied, assume now that we are given two versions $\theta_T$ and $\theta'_T$ of a NN, trained with two different training procedures up to iteration $T$. In case that $\theta_T$ and $\theta'_T$ are (significantly) different, then the adapted version of either Thm.\ \ref{thm:fNNfint} or \ref{thm:generaladaptive} shows that both models generally converge to different minimizers; this effect would persist even if the two NNs were trained by the same procedure for $t>T$. On the contrary, if $\theta_T$ and $\theta'_T$ were the same (or similar) at iteration $T$, then Thm.\ \ref{thm:sgdadaptiveinformal} suggests that the minimizers will nevertheless differ when $\theta_T$ is updated with adaptive (S)GD and $\theta'_T$ with vanilla (S)GD.
While our results only consider the impact on the minimizer obtained after an initial training period, there may exist further effects not considered in our analysis during the initial training steps $t<T$, where the features change.

	\begin{figure}[ht]
		\begin{center}
			\begin{minipage}[b]{0.49\textwidth}
				\includegraphics[width=\textwidth]{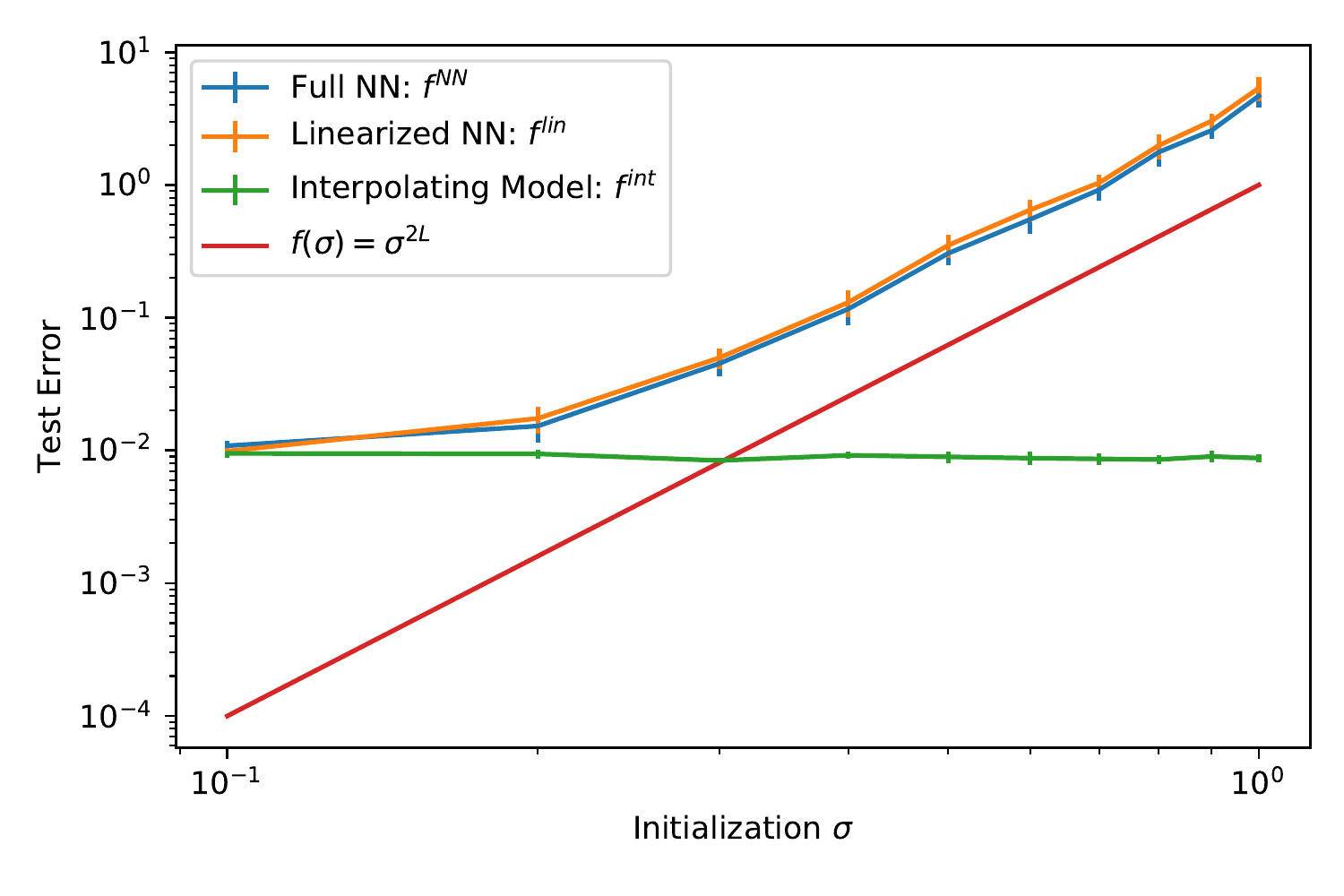}
			\end{minipage}
			\begin{minipage}[b]{0.49\textwidth}
				\includegraphics[width=\textwidth]{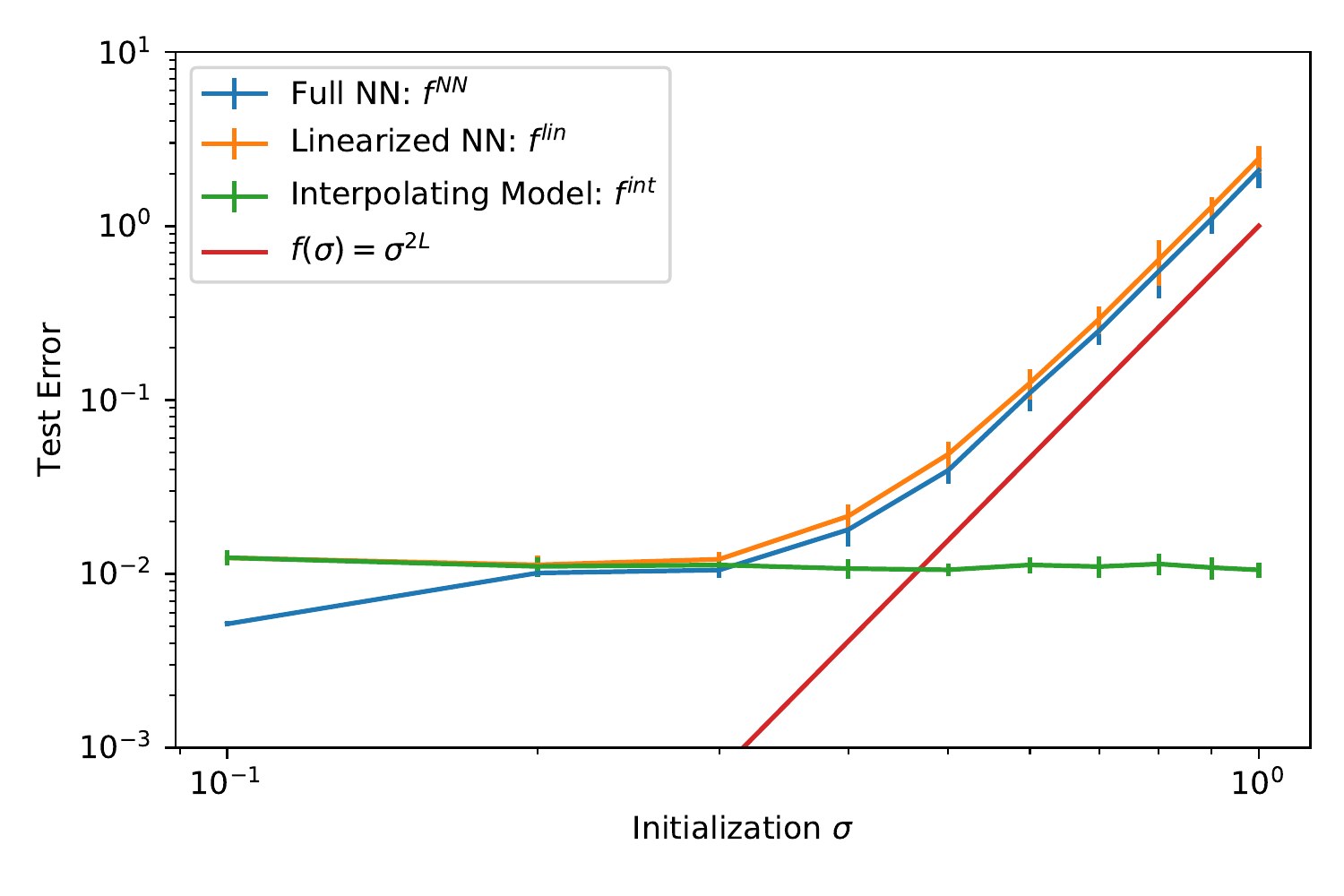}
			\end{minipage}
		\end{center}
		\caption{\textbf{Impact of initialization.} The impact of initialization on the test performance of trained overparameterized NNs is illustrated for $L=2$ (width $m=4000$, left) and $L=3$ layers ($m=2000$, right). At small initialization size $\sigma$, the trained $\fNN$ is close to the interpolating model $\fint$ (Sect.\ \ref{sect:init}), which is virtually independent of the initialization $\theta_0$ and $\sigma$. At larger $\sigma$, the test error grows strongly $\sim\sigma^{2L}$, with depth-dependent exponent (Thm.\ \ref{thm:initsigma}). All NNs were trained to the same low empirical error and are close to their linearizations $\flin$, verifying that indeed we are in the overparameterized limit. Our results underline that initialization does not merely provide a favorable starting point for optimization, but can strongly impact NN test performance.}
		\label{fig:init}
	\end{figure}
	
	\begin{figure}[ht]
	\begin{center}
		\begin{minipage}[b]{0.47\textwidth}
			\includegraphics[width=\textwidth]{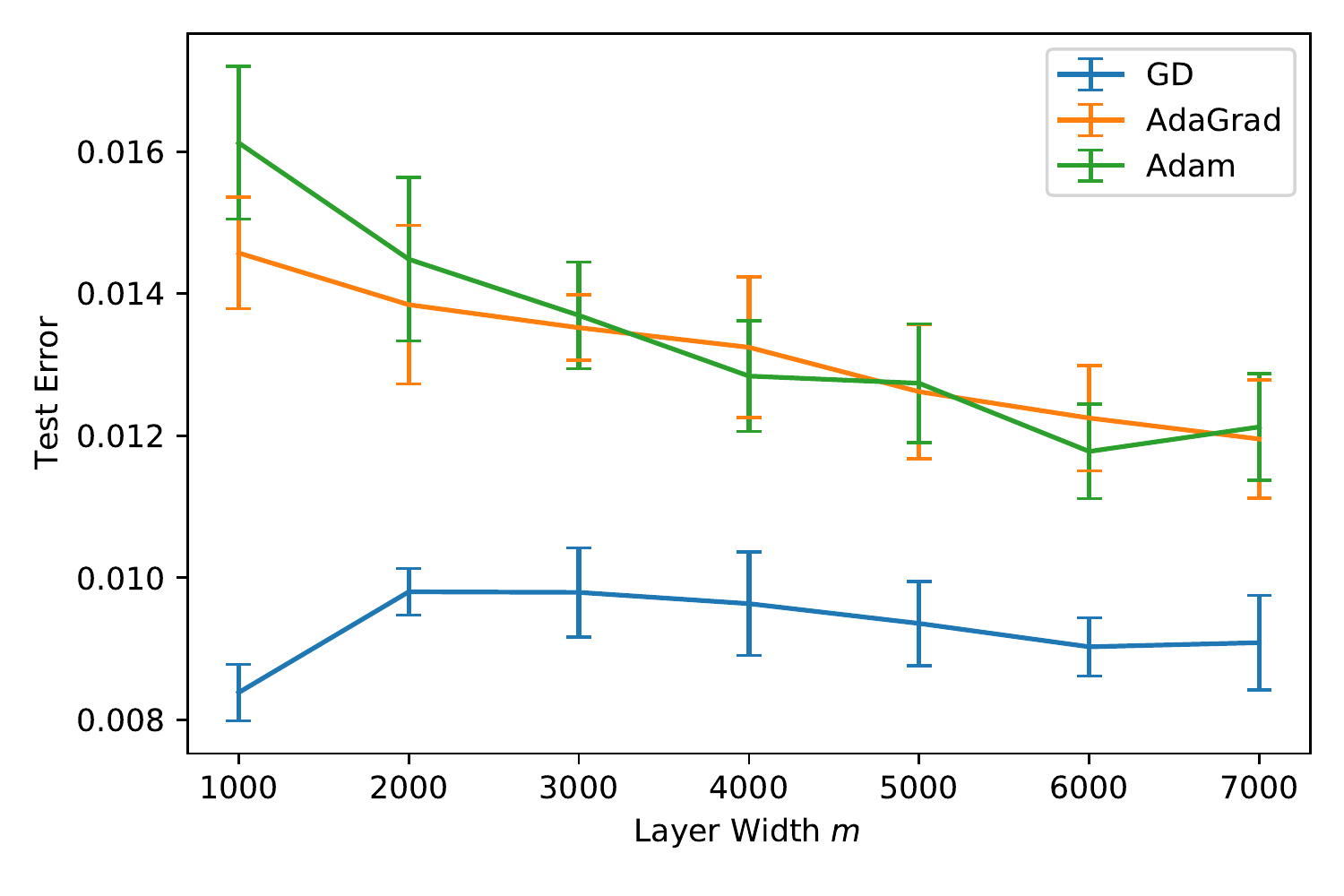}
		\end{minipage}
		\begin{minipage}[b]{0.47\textwidth}
			\includegraphics[width=\textwidth]{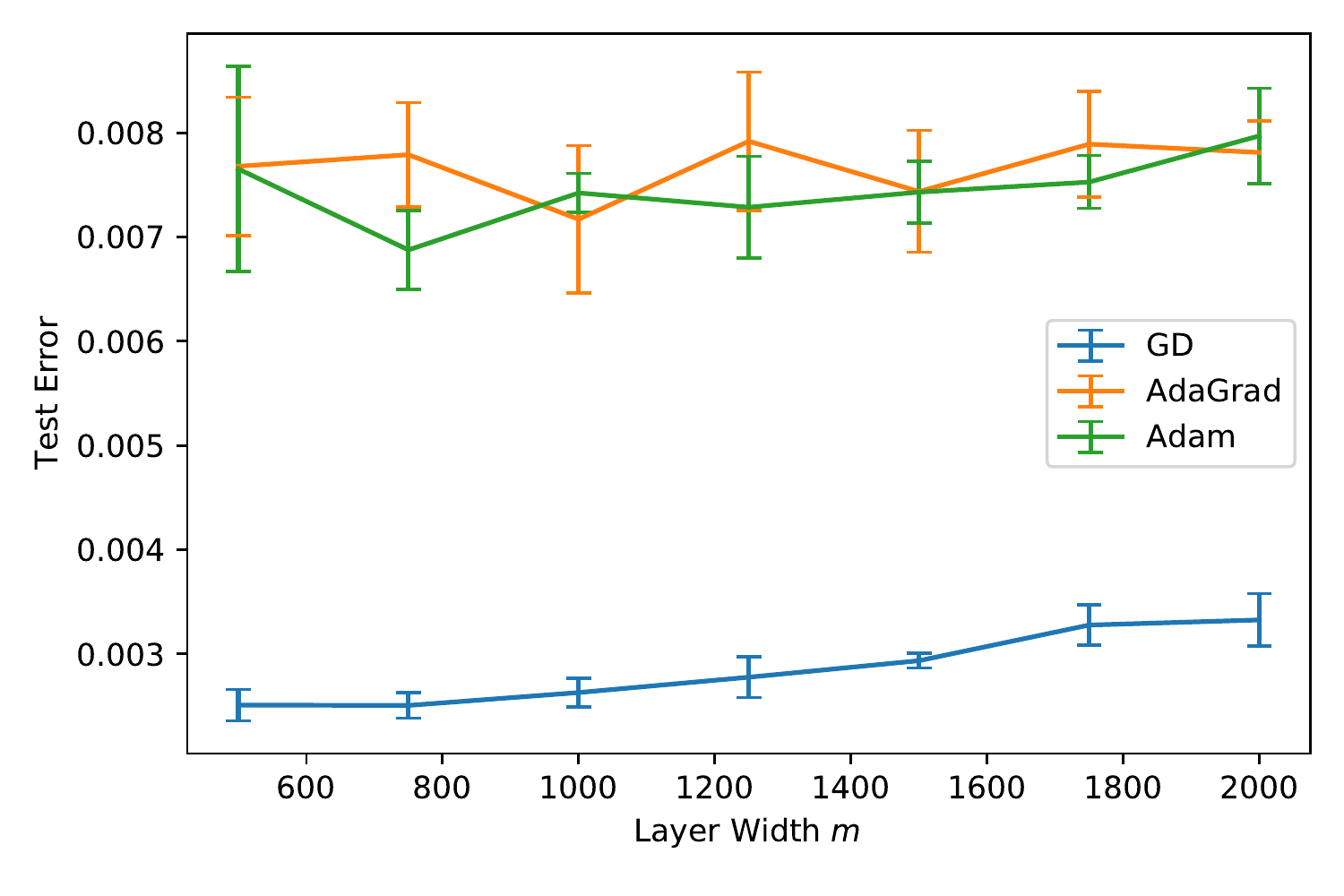}
		\end{minipage}
	\end{center}
	\caption{\textbf{Impact of adaptive optimization.} It is shown that the test performance depends strongly on the choice of adaptive optimization method, for a variety of overparameterized NNs with $L=2$ (left) and $L=3$ layers (right) over different widths $m$. Even though all models were trained to the same low empirical error, there is a significant performance gap between vanilla GD and the adaptive methods, underlining the results of Sect.\ \ref{sect:adaptive}. Thus, while adaptive methods for overparameterized NNs may improve the training dynamics, they also affect the minimizers obtained.}
	\label{fig:adaptive}
\end{figure}

	\section{Experiments}\label{sect:experiments}
\begin{figure}[ht]
		\begin{center}
			\begin{minipage}[b]{0.47\textwidth}
				\includegraphics[width=\textwidth]{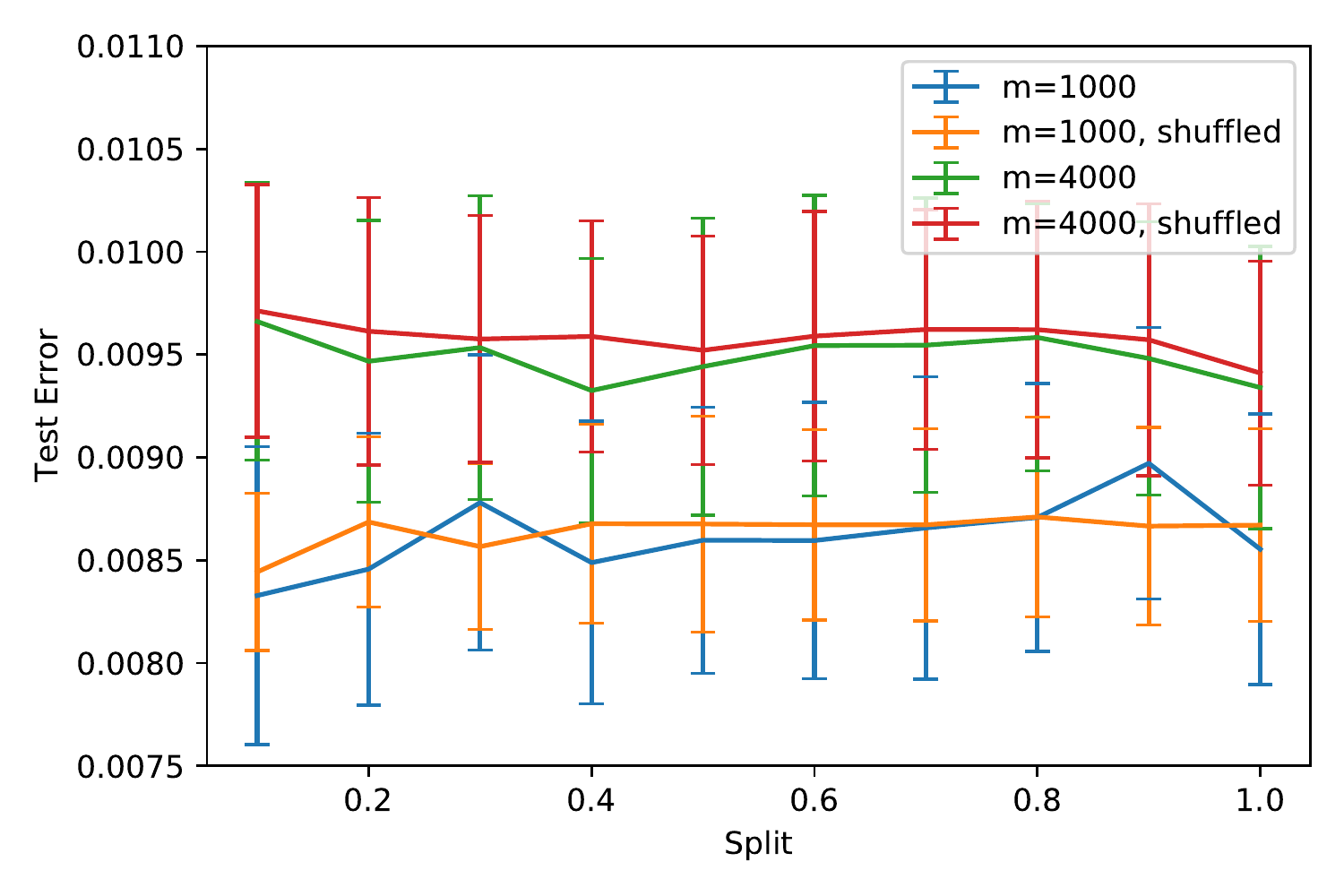}
			\end{minipage}
			\begin{minipage}[b]{0.47\textwidth}
				\includegraphics[width=\textwidth]{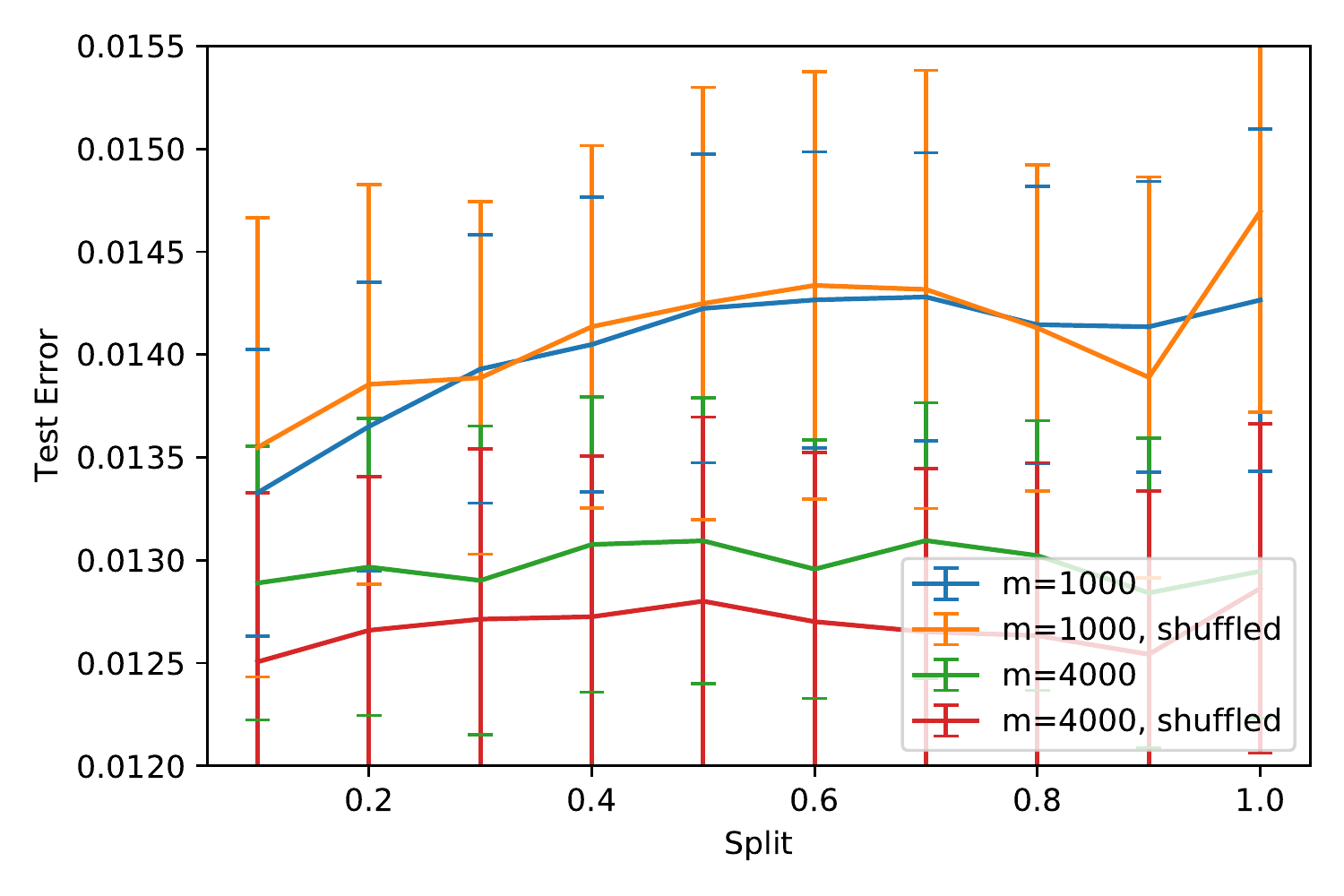}
			\end{minipage}
		\end{center}
		\caption{\textbf{Impact of stochastic training.} Without adaptive optimization (left), SGD-trained NNs have basically the same test error as GD-trained ones ($\text{split}=1.0$), for a large range of mini-batch sizes and both with and without shuffling the training set (Thm.\ \ref{thm:sgdadaptiveinformal}(a)). In contrast to this, when using adaptive optimization methods (AdaGrad, right) the test-performance becomes dependent on the batch-size (Thm.\ \ref{thm:sgdadaptiveinformal}(b)). The mini-batch size is given as the split ratio of the training set size, with vanilla GD corresponding to $1.0$. The NNs shown have $L=2$ layers and widths $m$ as indicated.}
		\label{fig:sgd}
	\end{figure}

\begin{figure}[ht]
		\begin{center}
			\begin{minipage}[b]{0.515\textwidth}
				\includegraphics[width=\textwidth]{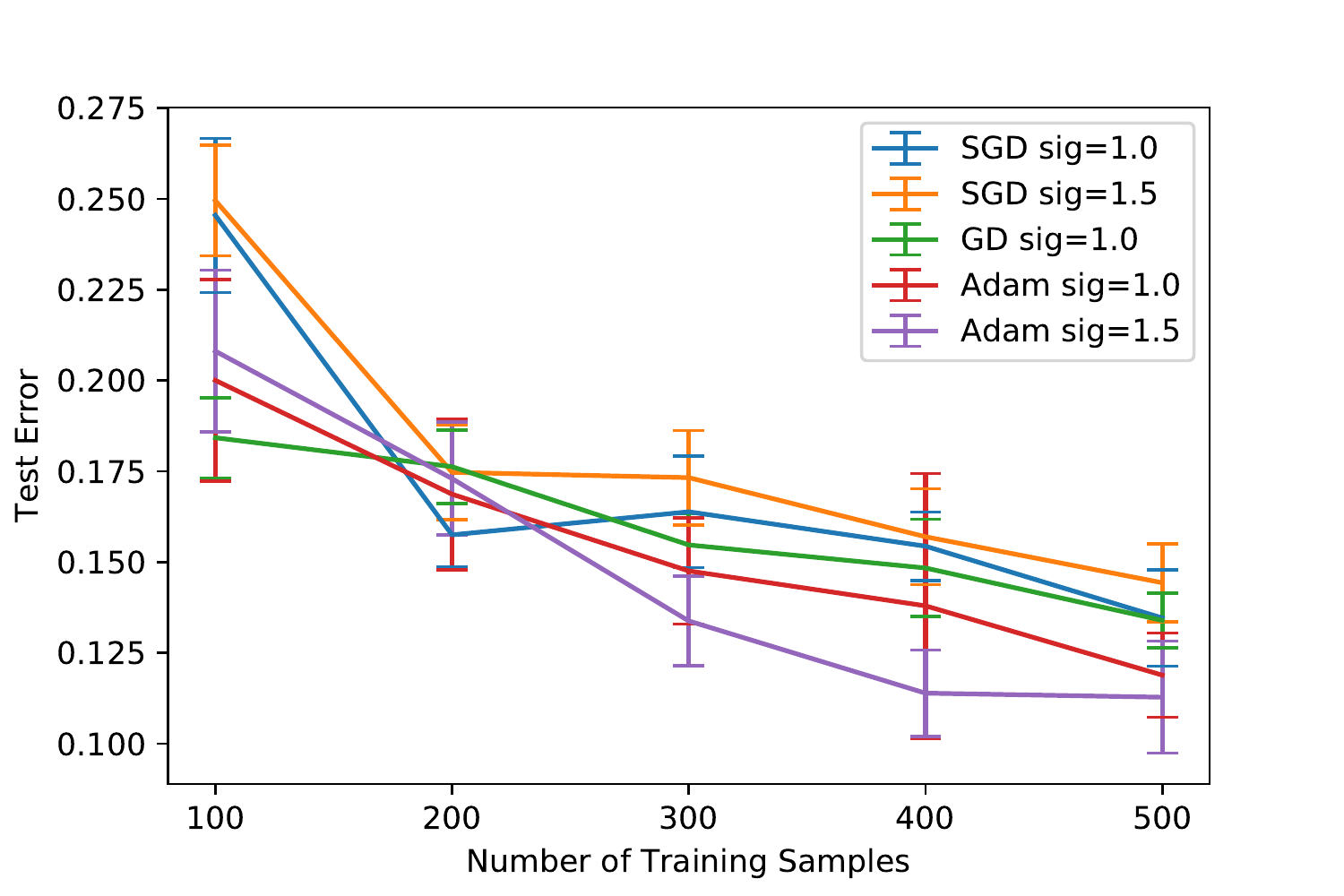}
			\end{minipage}
					\begin{minipage}[b]{0.47\textwidth}
			\includegraphics[width=\textwidth]{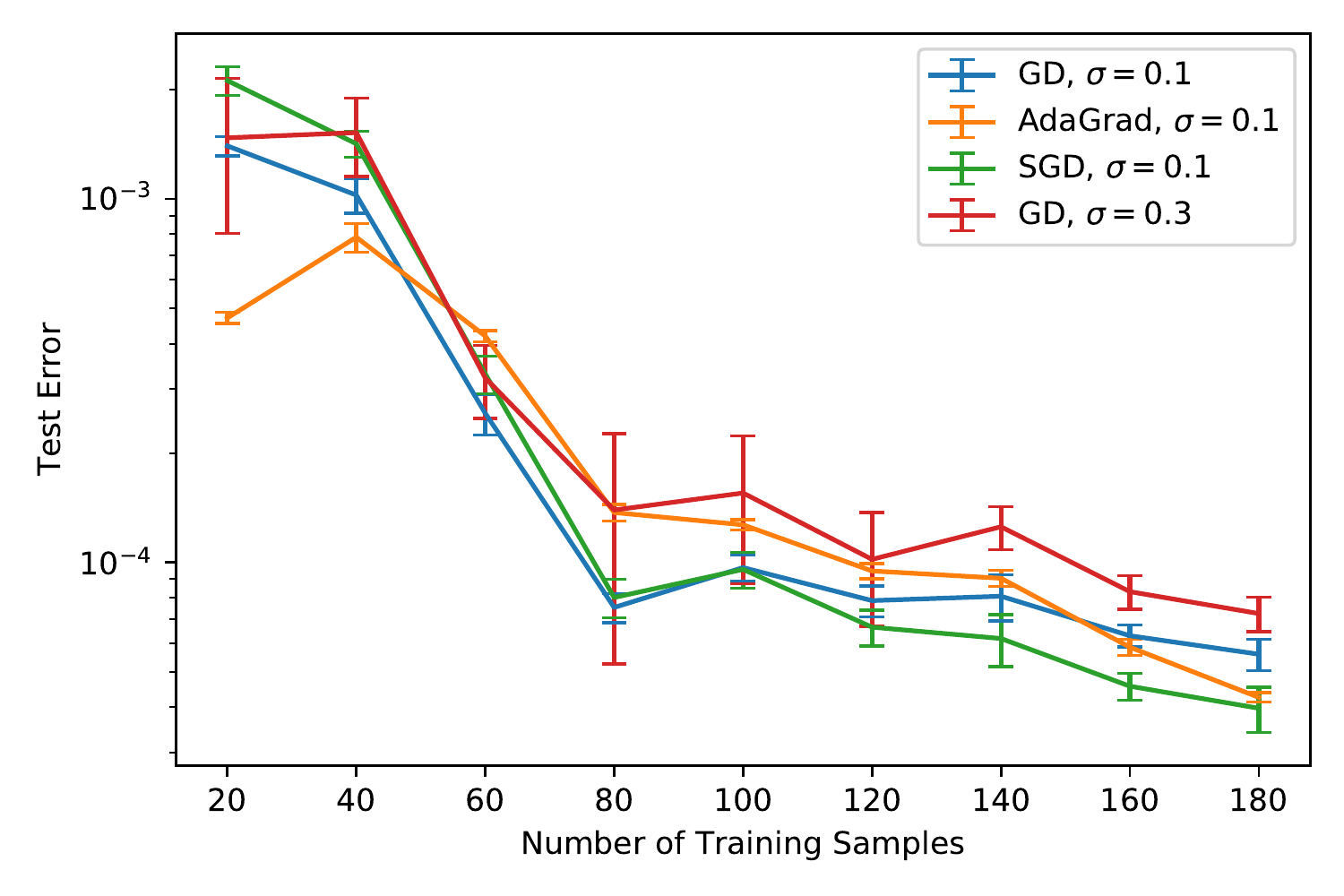}
		\end{minipage}
		\end{center}
		\caption{\textbf{ResNet networks and cross-entropy loss.} The left figure shows the test performance of ResNet-20 on the Cifar10 data set using ``airplaine'' vs.\ ``automobile'' labels (encoded as 0 and 1) trained with squared loss, for different optimization methods (GD, SGD, AdaGrad) and initialization sizes $\sigma$, over the training set size. The figure on the right shows the same setting as in Fig.\ \ref{fig:intro} but instead of the squared loss the network is trained using the cross-entropy loss. The shown results are averaged over 5 repetitions for each setting and all models are trained to the same low empirical error of $10^{-5}$. }
		\label{fig:modern}
	\end{figure}

	Here we demonstrate our theoretical findings experimentally. We perform experiments on three different data sets, MNIST (here), Fashion-MNIST (App.\ \ref{app:additional_experiments}), and CIFAR10 (Fig.\ \ref{fig:modern}), using their 0 and 1 labels. The first experiment (Fig.\ \ref{fig:init}) investigates the effect which the initialization size can have on the test performance of NNs, confirming the results of Sect.\ \ref{sect:init} qualitatively and quantitatively. Additionally, the behavior of the test (validation) error with $\sigma$ demonstrates the effectiveness of the error mitigation algorithm described in Sect.\ \ref{sect:init}.
The second experiment (Fig.\ \ref{fig:adaptive}) demonstrates the significant difference in test performance between NNs trained with vanilla GD and the adaptive optimization methods AdaGrad and Adam (Sect.\ \ref{sect:adaptive}). The third experiment (Fig.\ \ref{fig:sgd}) illustrates that for non-adaptive SGD there is only weak dependency on the batch-size and ordering of the datapoints, whereas for adaptive optimization with mini-batch SGD the dependence is noticeable (Sect.\ \ref{sect:sgd}).
	
These first three experiments are run with one and two hidden-layer NNs of different widths $m$. In line with our framework and with other works on overparameterized NNs we minimize the weights of the NN w.r.t.\ the empirical squared loss on a reduced number of training samples ($N=100$), to make sure that overparameterization $m\gg N$ is satisfied to a high degree. We train all the NNs to a very low training error ($<10^{-5}$) and then compare the mean test error from 10 independently initialized NNs with error bars representing the standard deviation\footnote{More details on the settings needed to reproduce the experiments can be found in App.\ \ref{app:experimental_setup}.}.

In addition to the effects of the three settings described in the previous paragraph, Fig.\ \ref{fig:intro} illustrates that similar effects appear in less overparameterized settings as well. Furthermore, while some of the theoretical conditions are not satisfied for modern architectures such as ResNets or networks trained with the cross-entropy loss, Fig.\ \ref{fig:modern} shows that comparable effects appear in these settings as well.

		\section{Related work}
	The fact that NN initialization can influence the average test performance was pointed out in \cite{zhang2019type} and  \cite{zhang2019identity}, who investigated this experimentally, but did not quantify it. The method suggested by \cite{zhang2019type} to reduce this effect doubles the number of NN parameters, significantly increasing the computational cost. Another method to reduce this effect was suggested by \cite{chizat2019lazy} with the ``doubling trick'', which was shown to potentially harm the training dynamics \cite{zhang2019type}. \cite{xiao2020disentangling} also investigates the impact of different initialization scalings but require deep and wide NNs, while we only requires wide NNs.
	Furthermore, \cite{wilson2017marginal} and \cite{zhang2019identity} observed a significant test performance gap between GD and adaptive optimization methods for overparameterized NNs. While \cite{wilson2017marginal} does provide an analytical toy example for this in a linear model, our analysis of NNs is general and holds for generic training data. \cite{shah2018minimum} attempts to explain such gaps for overparameterized linear models, but lacks the explicit expressions Eqs.\ (\ref{general_adaptive_formula}), (\ref{adagrad_solution}) we find. \cite{qian2019implicit} looks into the implicit bias of AdaGrad only for the cross-entropy loss and \cite{amari2021when} investigates the generalization behaviour of natural gradient descent under noise. While we on the other hand investigate the impact of general adaptive training methods for the squared loss on the minimizer obtained.

	The convergence of the training error under SGD in comparison to GD has been the subject of many works, recently also in the context of strongly overparameterized NNs \cite{ma2018power,sankararaman2019impact,allen2018convergence,borovykh2019effects}. We, on the other hand, investigate whether the minimizer found by SGD is similar to the GD minimizer on a test set. 	Another concept closely linked to this is the implicit bias of (S)GD on the trained NN, certain aspects of which have been investigated in \cite{soudry2018implicit,glorot2010understanding,rahaman2018spectral,oymak2018overparameterized,bietti2019inductive}. Our work elucidates the implicit bias caused by the random NN initialization and by the optimizer.
	
	We use interpolating kernel methods \cite{belkin2018understand} and the idea of an overparameterized learning regime which does not suffer from overfitting \cite{ma2018power,belkin2018does,belkin2018overfitting}. Bounds on their test performance have been derived by \cite{liang2018just} and \cite{arora2019fine}. \cite{arora2019exact} on the other hand demonstrates their good performance experimentally, but does not investigate how the scale of NN initialization or nonzero NN biases influence the test behavior (our Sect.\ \ref{sect:init}). While \cite{vaswani2020each} also explores different minimum norm solutions they only consider linear models and different loss functions and do not focus on the impact of different training techniques as we do. 
	
	Our analysis uses the Neural Tangent Kernel (NTK) limit \cite{jacot2018neural}, where the NN width scales polynomially with the training set size. In this regime, it is found that NNs converge to arbitrarily small training error under (S)GD. The first works to investigate this were \cite{Daniely2017Conj}, \cite{li2018learning}  and \cite{du2018gradient}. Later, \cite{allen2018convergence}, \cite{allen2018convergencernn}, and \cite{dudeep2018gradient} extended these results to deep NNs, CNNs and RNNs. The recent contribution by \cite{lee2019wide}, building upon \cite{lee2017deep} and \cite{matthews2018gaussian}, explicitly solves the training dynamics.
	
	\section{Discussion}
	We have explained theoretically how common choices in the training procedure of overparameterized NNs leave their footprint in the obtained minimizer and in the resulting test performance. These theoretical results provide an explanation for previous experimental observations \cite{wilson2017marginal,zhang2019identity}, and are further confirmed in our dedicated experiments.
	
	To identify and reduce the harmful influence of the initialization on the NN test performance, we suggest a new algorithm motivated by the bounds in Thm.\ \ref{thm:initsigma}. The potentially harmful influence of adaptive optimization on the test performance, however, cannot be reduced so easily if one wants to keep the beneficial effects on training time. Indeed, current adaptive methods experimentally seem to have worse test error than SGD (\cite{wilson2017marginal} and Fig.\ \ref{fig:adaptive}). Therefore, adaptive methods for overparameterized models should be designed not only to improve convergence to any minimum, but this minimum should also be analyzed form the perspective of statistical learning theory.

While our theory applies to NNs trained with squared loss, we believe the same effects to appear in NNs trained with cross-entropy loss (see Fig.\ \ref{fig:modern}). And while we show that in the less overparameterized regime, the minimizer is still dependent on training choices, it remains an open question to disentangle how exactly the learned data-dependent features contribute to the effects explained in our work.

\bibliographystyle{splncs04}
\bibliography{ECML_bib}

\newpage
\appendix
\begin{center}

    {\Large{Supplementary Material for}}
    
    \medskip
    
    {\LARGE{\textbf{Which Minimizer Does My}}}

\smallskip
    
    {\LARGE{\textbf{Neural Network Converge To?}}}

\end{center}

\smallskip
\section{Effects of initialization and adaptive training on underparameterized models}\label{app:underpara_model}
In this section we want to show that the choice of initialization and adaptive training have no effect on underparameterized NNs using the same techniques as we applied for strongly overparameterized NNs in this work. In the underparameterized regime the linearized NN does not stay close to the full NN during training. Therefore, we only investigate the linearized underparameterized NN $\flin_\theta(x)~:=~\fNN_{\theta_0}(x)+\phi(x)(\theta-\theta_0) $, where again $\phi(x):= \nabla_{\theta}\fNN_{\theta}(x)\big|_{\theta=\theta_0} \in \mathbb{R}^{1\times P}$ represents the feature vector. 

We want to demonstrate that in the underparameterized case the peculiar effects of Sect.\ \ref{sect:init} and \ref{sect:adaptive} do not persist. In the underparameterized regime all models, independent of the training method, converge to the same unique minimizer. This is also what one would expect, since the linear underparameterized model is strictly convex and thus the minimizer needs to be unique. When the number of parameters $P$ exceeds the number of training samples $N$ the models stays convex, but not strictly convex, since there are several weight configurations that all achieve zero training loss and thus there exist multiple minimizers.

\textbf{Initialization:} In the underparameterized regime the setup is almost identical to the one in Thm.\ \ref{thm:fNNfint} but instead of the NN output we use the linearized NN output and we have more data points than parameters $\left(P<N\right)$. Thus, the inverse of $\phi(X)\phi(X)^T$ does not exist with high probability over the initialization. Instead, for strongly underparameterized models, the inverse of $(\phi(X)^T\phi(X))$ exists with high probability over the initialization. Therefore, assuming that $(\phi(X)^T\phi(X))^{-1}$ exists and using the same approach and setting as in Thm.\ \ref{thm:fNNfint} for the underparameterized NN, we get for the fully converged linearized model

\begin{align*}
\flin(x) &= \phi(x)(\phi(X)^T\phi(X))^{-1}\phi(X)^T Y  \\ &+  \phi(x)\left[\ii-(\phi(X)^T(\phi(X))^{-1}\phi(X)^T\phi(X)\right]\theta_0\\
&= \phi(x)(\phi(X)^T\phi(X))^{-1}\phi(X)^T Y, \qquad \forall x\in \mathbb{R}^d ,
\end{align*}
which is as expected independent of $\theta_0$. The assumption that $(\phi(X)^T\phi(X))^{-1}$ exists holds with high probability over the initialization if $N$ is much larger than $P$. For less underparameterized models by adding $\lambda$ times a squared regularization term on the weights one can guarantee that the inverse exists. This changes the minimizer to $\fNN(x) = \phi(x)(\phi(X)^T\phi(X)+\lambda \ii)^{-1}\phi(X)^T Y$, which is still independent of $\theta_0$. 

\textbf{Adaptive Optimization:} Further, using similar techniques as in Thm.\ \ref{thm:generaladaptive}, it is again easy to see that adaptive optimization (see Eqs.\ \ref{adaptive_update}) has no effect on the fully converged NN for underparameterized models. Again assuming that $(\phi(X)^T\phi(X))^{-1}$ exists and the underparameterized NN trained with an adaptive optimization converges, we get 
\begin{align*}
\flin(x) &= \lim_{t \to \infty}\phi(x)(\phi(X)^T\phi(X))^{-1}\left(\ii - \prod_{i=t-1}^{0}(\ii-\frac{\eta}{N}\phi(X)^T\phi(X)D_i)\right)\phi(X)^T \\ &\times\left(Y- \fNN_{\theta_0}(X)\right)+\fNN_{\theta_0}(X) \\
&= \phi(x)(\phi(X)^T\phi(X))^{-1}\phi(X)^T Y, \qquad \forall x\in \mathbb{R}^d ,
\end{align*}
which is independent of the adaptive matrices $D_t$. A similar formula was also derived in \cite{shah2018minimum}.
\section{Interpolating kernel methods}\label{app:interpolating_kernel}
In this section we want to make the solutions of minimum complexity interpolating kernel methods explicit and show how they relate to minimum 2-norm interpolating methods. We start by investigating the minimum 2-norm interpolating method.
\begin{lemma}\label{lemma:minimum2norm}
	Let $X \in \mathbb{R}^{N\times d} $ be a non-degenerate data matrix and $Y\in \mathbb{R}^{N}$ a label matrix of a regression problem. Further, let $\fint(x) = \phi(x)\theta_\infty$ be a linear model, where we assume $\phi(x)$ to be expressive enough such that $\exists \theta^*: \phi(X)\theta^* = Y$, and $\theta_\infty$ given by
	\begin{equation}
	\theta_\infty = \argmin_{\theta} \|\theta\|^2_2 \quad \text{ subject to~~}Y= \phi(X)\theta.
	\end{equation}
	Then $\theta_\infty$ is given by 
	\begin{equation}
	\theta_\infty = \phi(X)\left(\phi(X)\phi(X)^T\right)^{-1}Y
	\end{equation}
\end{lemma}

\textbf{Proof.}
To solve the minimization problem above we can find the minimum of the Lagrangian with respect to all weights and the Lagrange multipliers $\lambda$. The Lagrange function takes the following form
\begin{equation*}
\mathcal{L}(\theta,\lambda) = \frac{\left\lVert \theta \right\rVert_2^2}{2} -\lambda^T\left(\Phi(X)\theta - Y\right)
\end{equation*}
Taking the derivative with respect to $\theta$ and setting it to zero gives us $\theta^T = \lambda^T\Phi(X)$. Now plugging this into the Lagrangian we get
\begin{equation}\label{dual_lagrangian}
\mathcal{L}(\lambda) = -\frac{1}{2}\lambda^T \Phi(X)\Phi(X)^T \lambda + \lambda^TY
\end{equation}
Minimizing this term with respect to $\lambda$ we get $\lambda = \left( \Phi(X) \Phi(X)^T\right)^{-1}Y$. Inserting this relation we get
\begin{equation*}
\theta_{\infty} =  \Phi(X)^T \left(  \Phi(X) \Phi(X)^T\right)^{-1}Y,
\end{equation*}
\QED

Next, we show that the minimum 2-norm solution is equivalent to the one of the minimum complexity interpolater w.r.t.\ a kernel norm.
\begin{lemma}
	Let $X \in \mathbb{R}^{N\times d} $ be the non-degenerate data matrix and $Y\in \mathbb{R}^{N}$ the label matrix of a regression problem and let $\fint(x) = \phi(x)\theta_\infty$ be a linear model, where we assume $\phi(x)$ to be expressive enough such that $\exists \theta^*\in \mathbb{R}^d: \phi(X)\theta^* = Y$, and $\theta_\infty$ given by
	\begin{equation}
	\theta_\infty = \argmin_{\theta} \|\theta\|^2_2 \quad \text{ subject to~~}Y= \phi(X)\theta.
	\end{equation}
	Further, let $\mathcal{H}_K$ be the RKHS of the kernel $K(x,y)= \phi(x)\phi(y)^T$ and $\hat{f}$ given by
	\begin{equation}
	\hat{f} =\argmin_{f \in \mathcal{H}_K} \left\lVert \tilde{f} \right\rVert_{\mathcal{H}_K} \text{ subject to }Y= f(X).
	\label{formula:minumumcomplexityRKHS}
	\end{equation}
	
	Then, $\forall x^* \in \mathbb{R}^d$ we have $\fint(x^*) = \hat{f}(x^*)$.
\end{lemma}
\textbf{Proof.}
We start by solving for the minimum complexity interpolating kernel problem given in (\ref{formula:minumumcomplexityRKHS}) and then compare the result with the solution we obtained from lemma \ref{lemma:minimum2norm}.
We can find the solution of the minimum complexity interpolating kernel problem by minimizing the following Lagrangian:
\begin{equation*}
\mathcal{L}(f,\lambda) = \frac{1}{2} \left\lVert \tilde{f} \right\rVert_{\mathcal{H}_K} +\lambda^T \left(Y- f(X) \right)
\end{equation*}
Because the RKHS-Norm is bounded by definition (for non degenerate data)  we can use the representer theorem ($f= \sum_{i=1}^N \alpha_i K(x_i,.)$ with $\alpha_i \in \mathbb{R}$) and the reproducibility of the kernel $K$ to arrive at
\begin{equation*}
\mathcal{L}(\alpha,\lambda) = \frac{1}{2} \alpha^T K(X,X)\alpha  +\lambda^T \left(Y-  K(X,X) \alpha \right)
\end{equation*}
Now taking the derivative with respect to $\alpha$, we get $\alpha = \lambda$. Inserting this into our Lagrangian we get
\begin{equation*}
\mathcal{L}(\lambda) = -\frac{1}{2}\lambda^T K(X,X) \lambda+ \lambda^TY
\end{equation*}
This Lagrangian is equivalent to the Lagrangian of the minimal two norm solution in the dual representation (\ref{dual_lagrangian}) and thus the two problems lead to the same solution.
\QED

\section{Detailed discussion of the impact of stochastic optimization}\label{app:SGDdetails}
This section provides further details on Sect.\ \ref{sect:sgd} in the main paper and also the full version of Thm.\ \ref{thm:sgdadaptiveinformal}.
First, we present the full version of Thm.\ \ref{thm:sgdadaptiveinformal} a), which states that NNs trained with adaptive GD and mini-batch SGD, with constant adaptive matrices, converge to almost the same minimizer independent of the batch-size. We get
\begin{theorem}\label{thm:gd_sgd}
	Let $\fNN_{adGD}$ and $\fNN_{adSGD}$ be fully trained strongly overparameterized NNs (\ref{NNparametrization}) trained using adaptive GD and SGD, with arbitrary mini-batch size but constant adaptive matrices $D_t= const$, under empirical squared loss on a non-degenerate dataset $\mathcal D$, where the weights have been initialized to the same values. Then, for sufficiently small learning rate $\eta$ there exists $C= {\mathrm poly}(N,1/\delta,1/\lambda_0,1/\sigma)$ such that for NNs with $m \geq C$ and $\forall x \in B_1^d(0)$ it holds with probability at least $1-\delta$ over the random initialization that
	\begin{equation*}
	\big|\fNN_{adGD}(x)-\fNN_{adSGD}(x)\big|\leq O(1/\sqrt{m})
	\end{equation*}
\end{theorem}

The idea to prove this result is to first utilize the fact that SGD can be written using an adaptive update step (\ref{adaptive_update}) with adaptive matrices $D_{B_t}:= \frac{N}{|B_t|}\left(\phi(X_{B_t})\right)^T\left(\phi(X)\phi(X)^T\right)^{-1}\phi(X)$ to show that the two linearized models converge exactly to the same minimizer. Then, we again use that strongly overparameterized NNs stay close to their linearized model, which also holds for SGD. The full proof can be found in App.\ \ref{app:proof_generalsgd}.
The result holds not only for small learning rates but is easy to generalize to any model in the NTK-regime that converges to a minimizer with zero training loss.

The part b) of Thm.\ \ref{thm:sgdadaptiveinformal} shows that when combining mini-batch SGD with adaptive optimization methods, where the adaptive matrices $D_t$ change during training, the two minimizers not only differ due to the linearization error but also by a path dependent contribution. The full theorem takes the following form

\begin{theorem}\label{thm:sgdadaptive}
	Given a NN $\fNN_{\theta}$ (\ref{NNparametrization}) and a non-degenerate training set ${\mathcal D}=(X,Y)$ for adaptive mini-batch SGD training (\ref{generalsgdstep}) under squared loss with adaptive matrices $D_t \in{\mathbb{R}}^{P\times P}$ concentrated around some $D$, there exists $C= {\mathrm{poly}}(N,1/\delta,1/\lambda_0,1/\sigma)$ such that for any width $m \geq C$ of the NN and any $x \in B_1^d(0)$ it holds with probability at least $1-\delta$ over the random initialization that
	\begin{align}
	&\fNN_{\theta_t}(x)  =\phi(x)A_t\left[ \ii - \prod_{k=t-1}^{0}\big(\ii - \frac{\eta}{N}\phi(X) D_k D_{B_k}\phi(X)^T\big)\right] \left(Y-\fNN_{\theta_0}(X)\right) \nonumber\\ & \ \ \ \ \ \ \  \ \ \  +\fNN_{\theta_0}(x)+\phi(x)B_t  +O\Big(\frac{1}{\sqrt{m}}\Big), \nonumber\\
	&\text{where}\quad A_t ~=~ D_{t-1}\phi(X)^T\left(\phi(X) D_{t-1} \phi(X)^T\right)^{-1} 	\text{and}\\ 
&B_t ~=~ \sum_{v=2}^{t}(A_{v-1}\!-\!A_v)\left[\ii -\!\!\!\!\prod_{w=v-2}^{0}\big(\ii\!-\!\frac{\eta}{N} \phi(X)D_wD_{B_w}\phi(X)^T\big) \right]\left( Y-\fNN_{\theta_0}(X)\right).\nonumber
	\end{align}
	
\end{theorem}

The main idea to prove this theorem is to again first make use of the fact that adaptive SGD can be written using the adaptive update rule (\ref{adaptive_update}), where the new adaptive matrices are a product of $D_{B_t}$ and the original adaptive matrices $D_t$ as explained in the main text. 
Then, making use of the results of Thm.\ \ref{thm:gd_sgd} we see that the NN output can be expressed in a similar way as in Thm.\ \ref{thm:generaladaptive}, but where now $A_{t} =D_t\phi(X)^T\left(\phi(X)D_t\phi(X)^T\right)^{-1}$ is independent of the SGD part of the adaptive matrices. Thus, for $t\to \infty$ the dependency on the $D_{B_t}$ matrices vanishes in the interpolating term and only contributes to the path dependent term $B_t$.

\section{Experimental setup}\label{app:experimental_setup}
\label{experiments}
In this section we want to give further details on how we conducted the experiments. Since the experimental settings are not complicated this should be more than enough to reproduce our results. All our experiments were run using the Pytorch package.

As described in the experimental section (Sect.\ \ref{sect:experiments}) we use three different experimental settings to investigate the dependence of the minimizer the NNs converge to on the initialization, the adaptive optimization method, and the use of SGD. The initialization scheme and the specific parametrization of our NNs are described in the notation section (see Sect.\ \ref{sec:notation}). For the learning rate we make either use of the theoretical value of the linear model and set the learning rate to $\eta = \frac{0.5}{\lambda_{max}(\phi(X)\phi(X)^T)}$ or, for the adaptive methods, we choose the largest value for the learning rate that still guarantees a monotonous decay of the training loss. In all experiments we use the ReLU-activation functions. For both datasets MNIST and Fashion-MNIST \footnote{Dataset: yann.lecun.com/exdb/mnist/ and github.com/zalandoresearch/fashion-mnist} we only use a reduced number of training samples ($N=100$) to ensure that we are in the strongly overparameterized regime without the need for overly wide NNs. We train the NNs until the empirical loss first drops bellow $10^{-5}$. This training error is well much smaller than the test errors we achieve and is chosen in such a way that the test error does not change significantly if we were to continue training to an even lower empirical loss. We then compare the mean test performance over 10 independently initialized NNs, for the initialization and the adaptive training experiments, and use the same 10 initializations for each split of the SGD experiments. The test performance is determined using 100 test points.

\begin{figure}[ht]
	\begin{center}
		\begin{minipage}[b]{0.47\textwidth}
			\includegraphics[width=\textwidth]{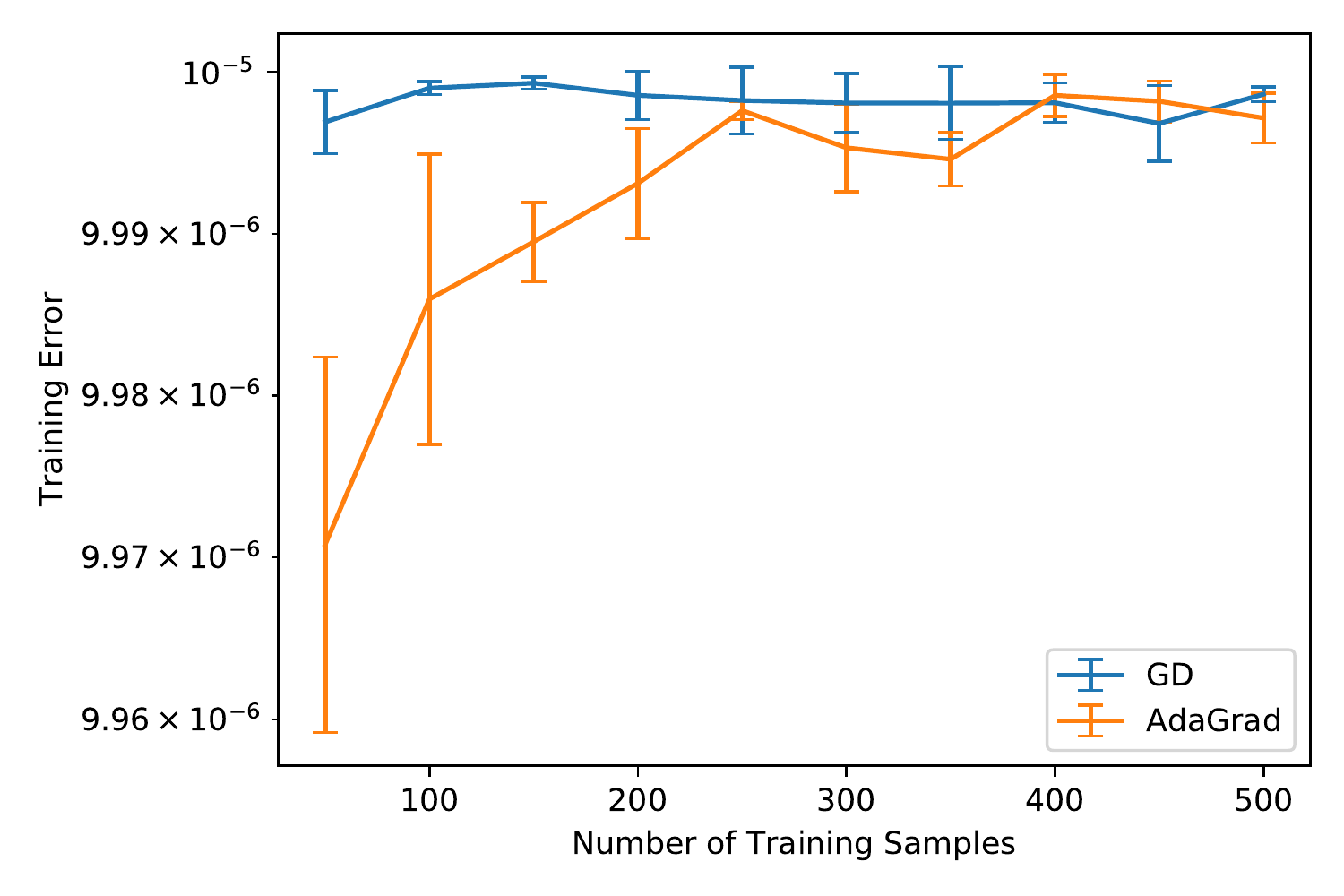}
		\end{minipage}
	\end{center}
	\caption{\textbf{Final Training error.} This figure shows the mean and standard error of the final training error when stopping after the training error first drops below $10^{-5}$. These results show that there is only a very  small difference in the training errors.}
	\label{fig:app_training_error}
\end{figure}

\newpage

\section{Additional experiments -- Fashion-MNIST and approximation}\label{app:additional_experiments}
\begin{figure}[ht]
	\begin{center}
		\begin{minipage}[b]{0.47\textwidth}
			\includegraphics[width=\textwidth]{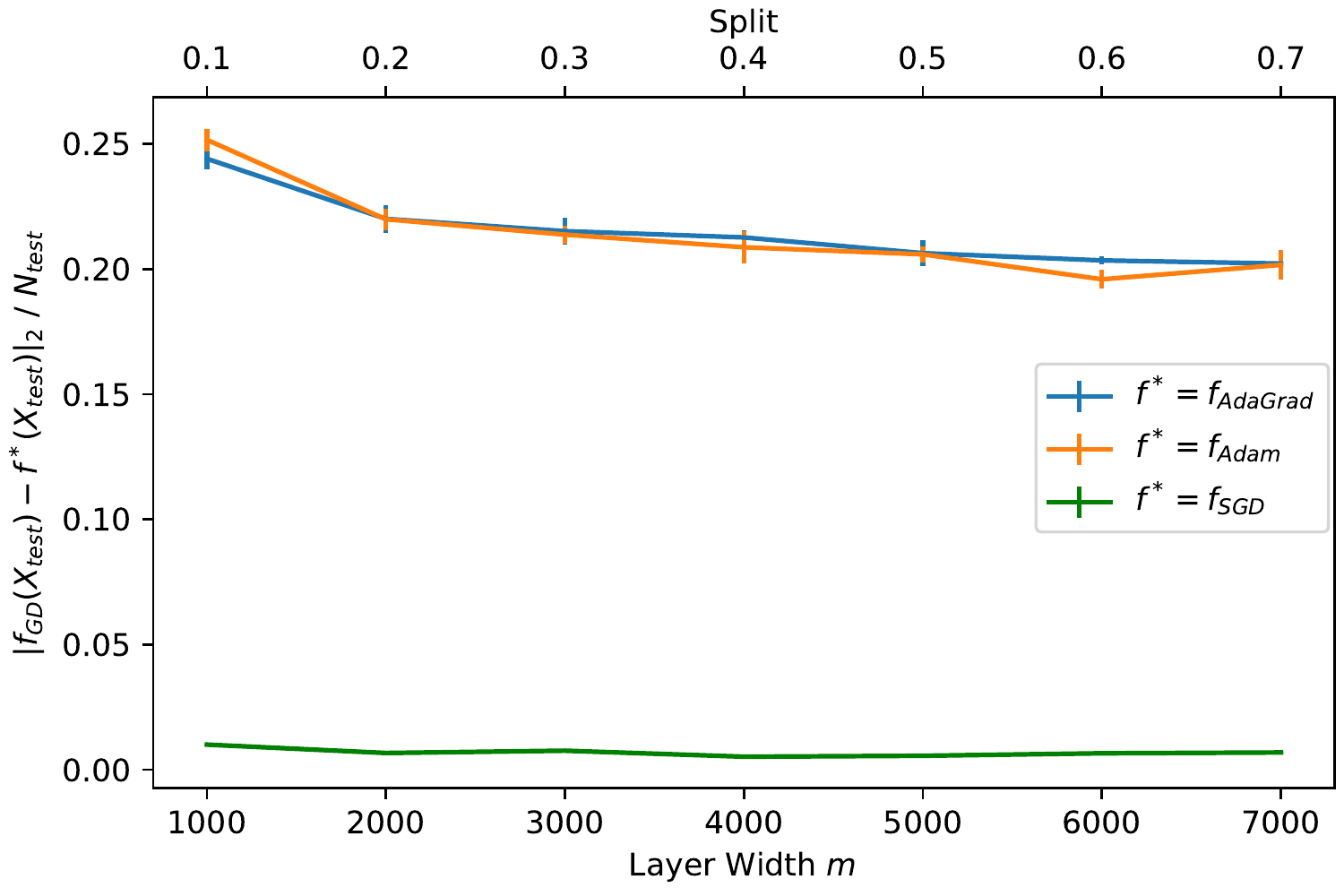}
		\end{minipage}
	\end{center}
	\caption{The figure shows the 2-norm of the difference for NN outputs of 100 test points of NNs trained with AdaGrad, Adam and SGD compared to plain GD on MNIST data. One can clearly see that the deviation imposed by SGD is very small compared to the changes due to adaptive training.}
	\label{fig:app_fnn_error}
\end{figure}

In this section we present the corresponding experimental results for the Fashion-MNIST dataset. The results show the same behaviour as the results in the experimental section (see Sect.\ \ref{sect:experiments}). Further,we show the actual deviation of the NN output trained with SGD, AdaGrad and Adam compared to the NN trained by vanilla GD (see Fig.\ \ref{fig:app_fnn_error}).
\begin{figure}[ht]
	\begin{center}
		\begin{minipage}[b]{0.47\textwidth}
			\includegraphics[width=\textwidth]{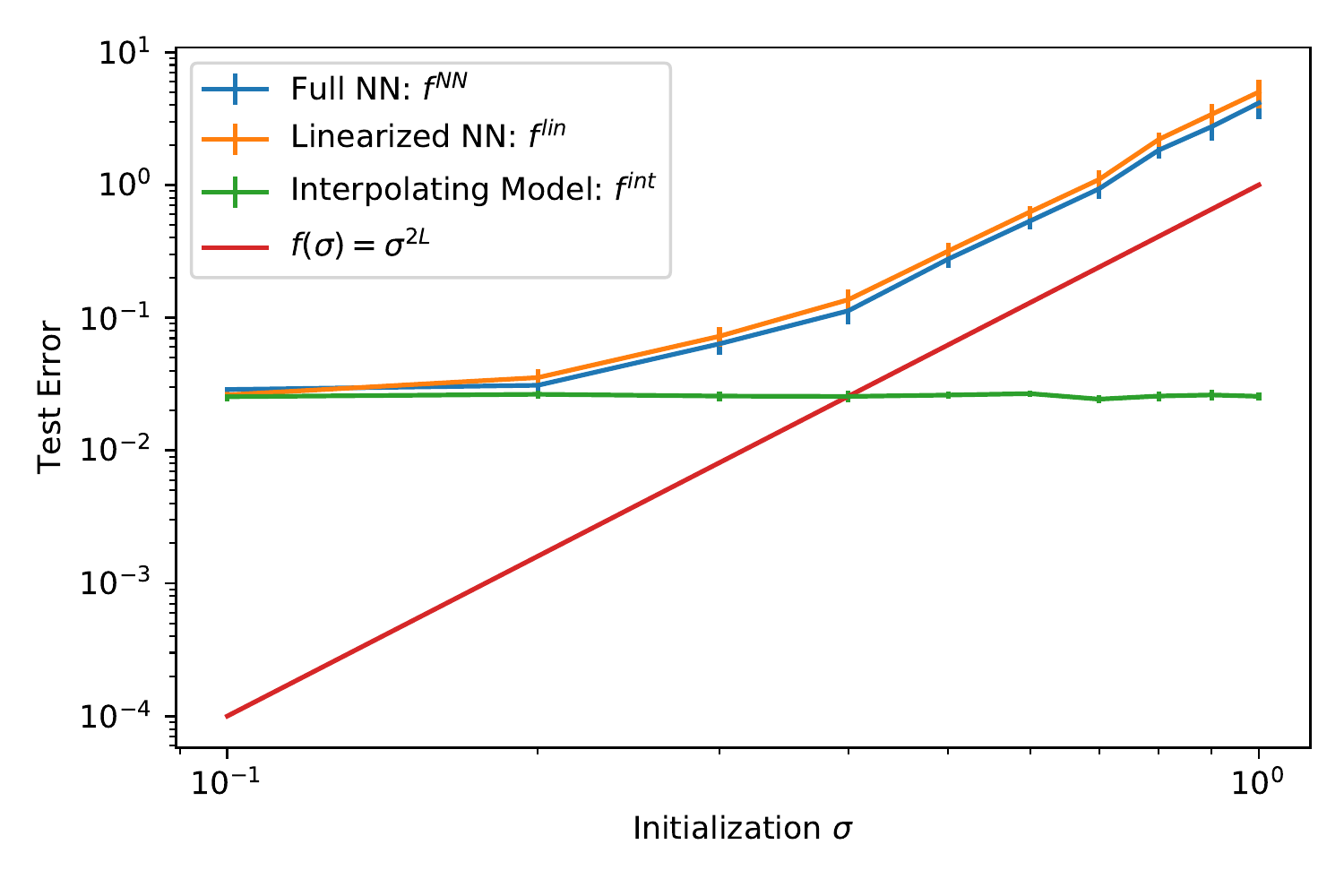}
		\end{minipage}
		\begin{minipage}[b]{0.47\textwidth}
			\includegraphics[width=\textwidth]{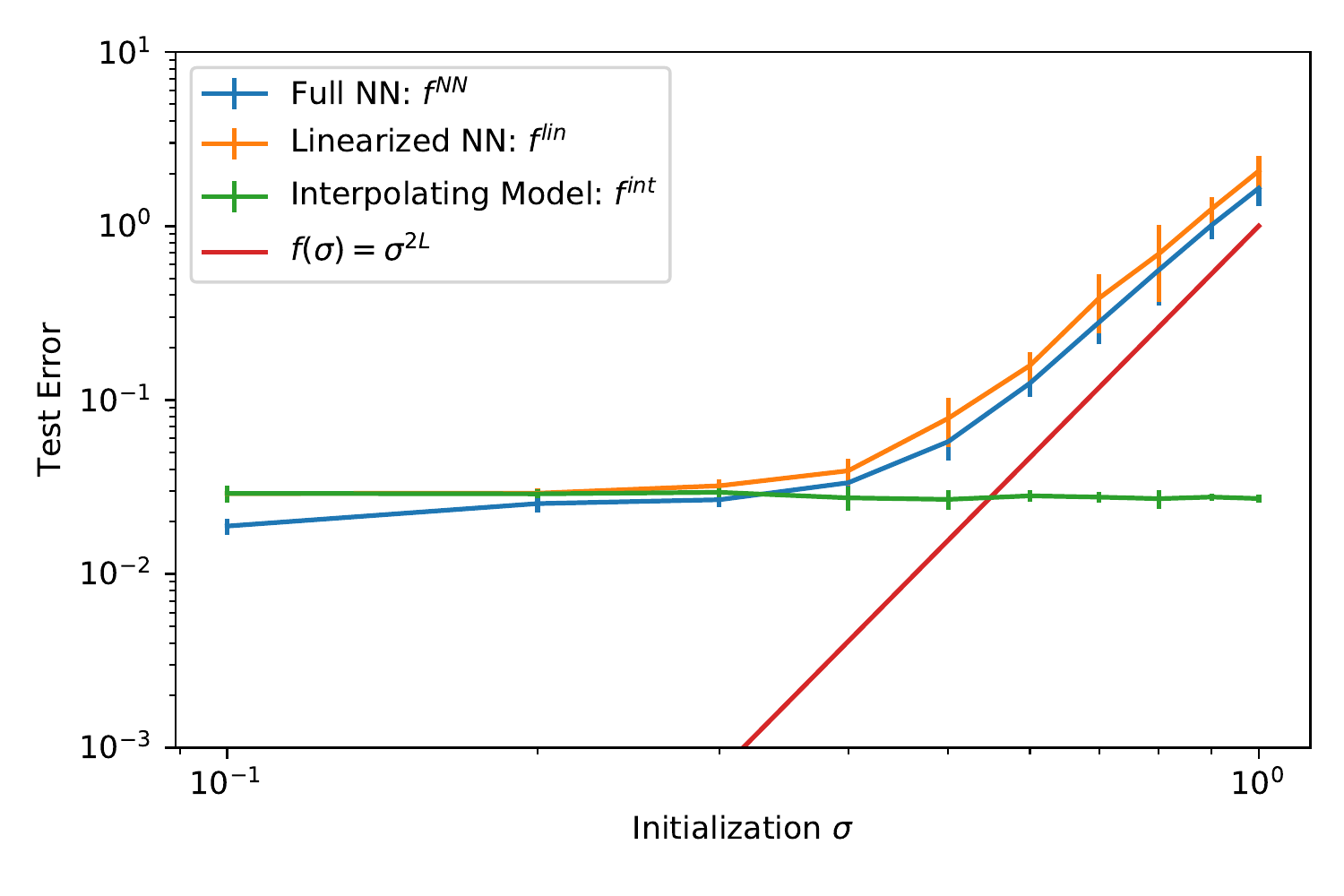}
		\end{minipage}
	\end{center}
	\caption{This figure shows the same setting as Fig.\ \ref{fig:init} but for Fashion-MNIST. The results are almost identical but with a larger test error since Fashion-MNIST is a more complex dataset.}
	\label{fig:app_init}
\end{figure}
\begin{figure}[ht]
	\begin{center}
		\begin{minipage}[b]{0.47\textwidth}
			\includegraphics[width=\textwidth]{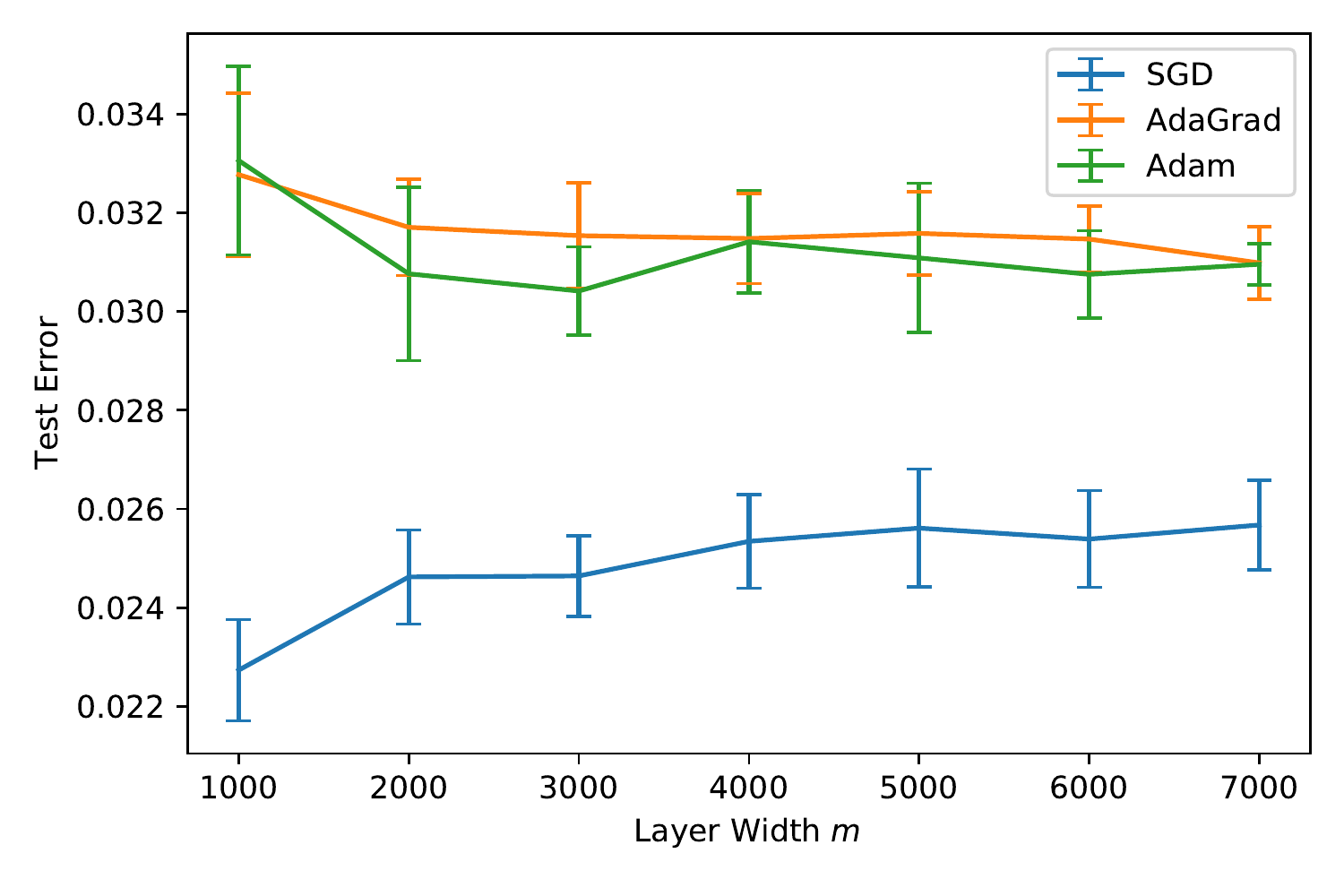}
		\end{minipage}
	\end{center}
	\caption{This figure shows the same setting as Fig.\ \ref{fig:adaptive} but for Fashion-MNIST. The result are almost identical but with a larger test error since Fashion-MNIST is a more complex dataset.}
	\label{fig:app_adaptive}
\end{figure}
\begin{figure}[ht]
	\begin{center}
		\begin{minipage}[b]{0.47\textwidth}
			\includegraphics[width=\textwidth]{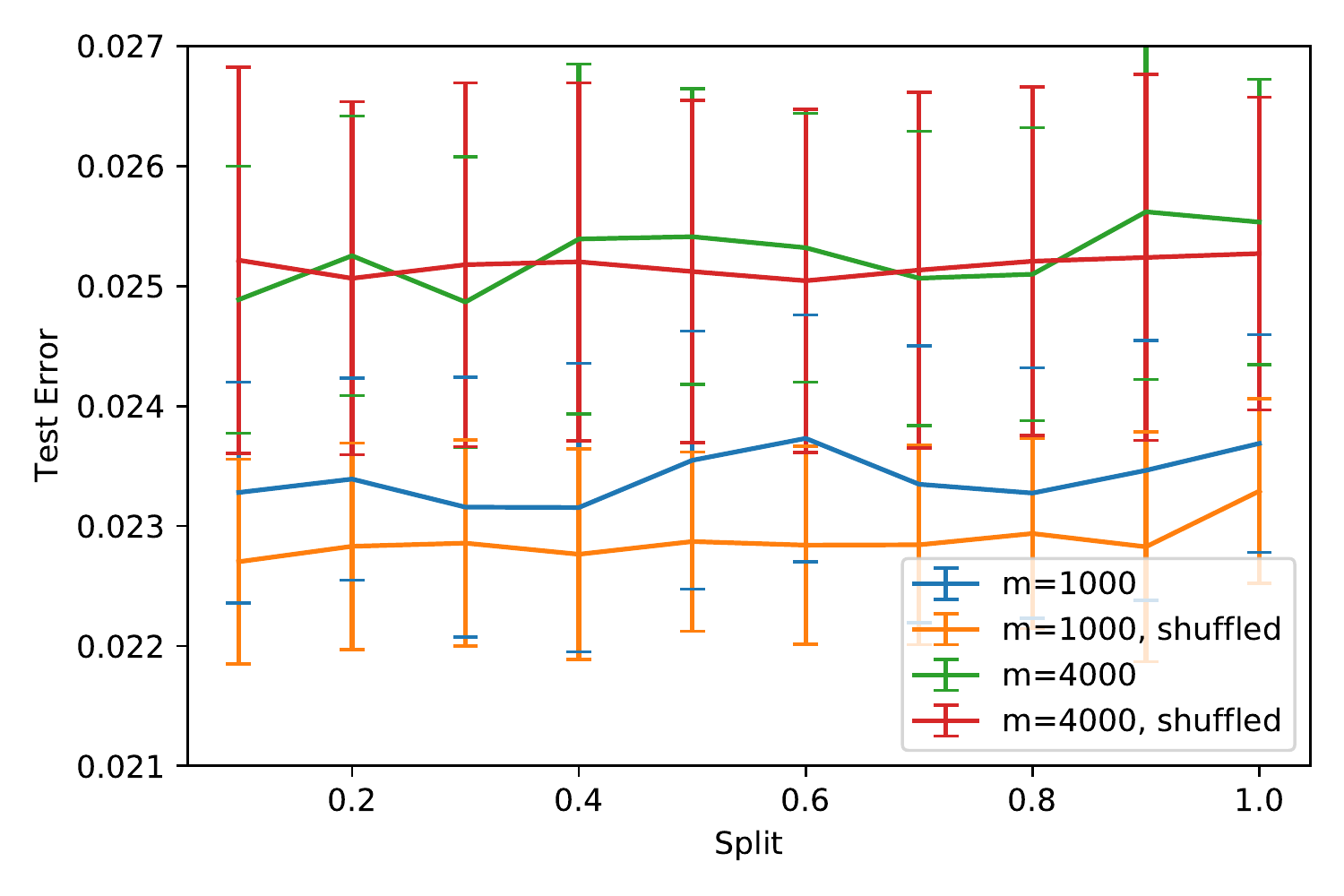}
		\end{minipage}
		\begin{minipage}[b]{0.47\textwidth}
			\includegraphics[width=\textwidth]{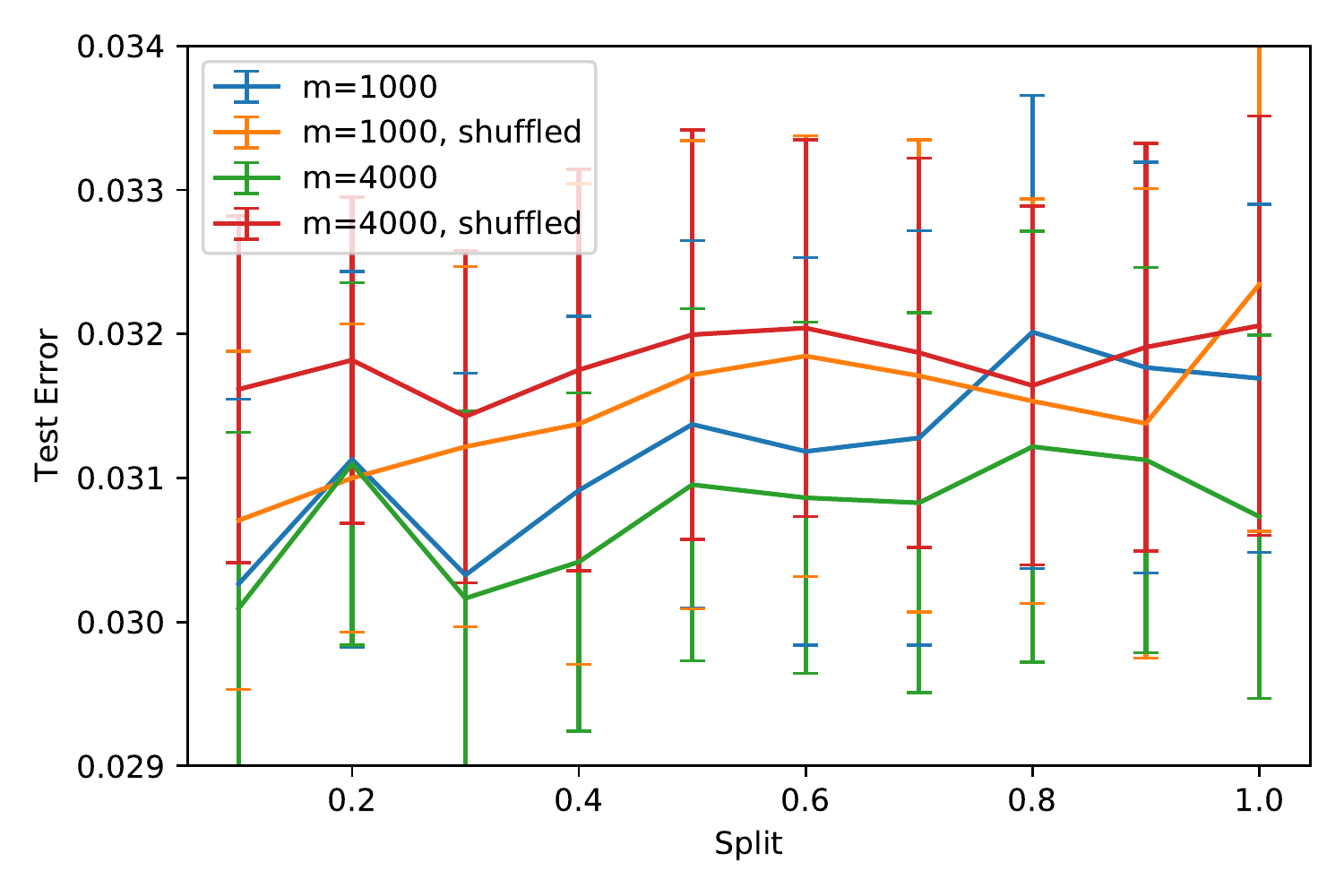}
		\end{minipage}
	\end{center}
	\caption{This figure shows the same setting as Fig.\ \ref{fig:sgd} but for Fashion-MNIST. The result are almost identical but with a larger test error since Fashion-MNIST is a more complex dataset.}
	\label{fig:app_sgd}
\end{figure}

\newpage

\section{Proofs and technical lemmas}\label{app:proofs}
\subsection{Proof of Thm.\ \ref{thm:fNNfint}}\label{app:proof_fNNfint}
\textbf{Proof.}
We start by analyzing the training behaviour of the linearized model and then show that the linear training dynamics is w.h.p. over initializations close to the training dynamics of the full NN in the strongly overparameterized regime.
The linear approximation of $\fNN$ is defined as
\begin{align*}
\flin_\theta(x)&:=\fNN_{\theta_0}(x)+(\nabla_\theta\fNN_\theta(x)\big|_{\theta_0})(\theta-\theta_0) \\
&= \fNN_{\theta_0}(x)+\phi(x)(\theta-\theta_0)
\end{align*}
For the weight updates of the linearized model with GD with respect to the empirical squared loss and learning rate $\eta$ we get
\begin{align*}
\theta_{t+1} &= \theta_t- \frac{\eta}{N}\phi(X)^T\left(\phi(X)(\theta_t-\theta_0)+\fNN_{\theta_0}(X)-Y\right) \\
&= \theta_t- \frac{\eta}{N}\phi(X)^T\left(\phi(X)\theta_t-\hat{Y}\right),
\end{align*}
where $\hat{Y} = Y+\phi(X)\theta_0- \fNN_{\theta_0}(X)$.
This iteration can be solved explicitly using an induction and the binomial theorem.
\begin{equation}
\begin{aligned}
\theta_{t} &= \theta_0+\phi(X)^T\left( \sum_{i= 1}^{t} (-1)^{i-1} \binom{t}{i}\left(\frac{\eta}{N}\right)^i (\phi(X)\phi(X)^T)^{i-1}\right)\left(\hat{Y}-\phi(X)\theta_0\right) \\
&= \theta_0+\phi(X)^T(\phi(X)\phi(X)^T)^{-1}\left[\ii-\left(\ii-\frac{\eta}{N} \phi(X) \phi(X)^T\right)^t\right] \left(\hat{Y} -\phi(X)\theta_0\right).
\end{aligned}
\end{equation}
If the learning rate satisfies $\eta < 2N/\lambda_{\text{max}}(\phi(X)\phi(X)^T)$ such that gradient descent converges, we get $\left(\ii-\frac{\eta}{N} \phi(X) \phi(X)^T\right)^t \to 0$ in the limit $t \to \infty$ as the spectral radius is smaller than one and the term in square brackets can be simplified to $\ii$. For this to hold we need  $\lambda_{\text{min}}(\phi(X)\phi(X)^T) > 0 $. Now, since $\phi(X)\phi(X)^T \to \Theta(X,X)$ in contribution and $\lambda_0 >0$ for non-degenerative datasets there exists a $C_1={\mathrm poly}(N,1/\lambda_0,1/\delta,1/\sigma)$ such that with probability $1 -\delta/2$ over random initializations we have $\lambda_{\text{min}}(\phi(X)\phi(X)^T) > \lambda_0 /2 >0$ \cite{du2018gradient}.

Plugging the result in for $\theta_\infty$ ($t \to \infty$) into $\flin_\theta$ results in
\begin{equation}
\flin(x)= \fNN_{\theta_0}(x)+ \phi(x)\phi(X)^T(\phi(X)\phi(X)^T)^{-1} (Y-\fNN_{\theta_0}(X)).
\end{equation}
Now making use of Lem.\ \ref{lemma:reluexpression}, where it is shown for NNs with ReLU-activation function that $\fNN_{\theta}(x) =\frac{1}{L}\langle\theta,\phi(x)\rangle$ we get
\begin{align*}
\flin(x)  &= \phi(x) \phi(X)^T(\phi(X)\phi(X)^T)^{-1}Y \\ &+  \frac{1}{L} \phi(x)\left[\ii-\phi(X)^T(\phi(X)\phi(X)^T)^{-1} \phi(X)\right]\theta_0
\end{align*}
Next using Lem.\ \ref{lemma:lee2019wide} we see that there exists $C_2= {\mathrm poly}(N,1/\delta,1/\lambda_0,\sigma)$ such that for NNs with a number of hidden units $m$ with $m \geq C_2$ we get with high probability $1-\delta/2$ over initializations that
\begin{equation}
\fNN(x) = \flin(x) +O\left(\frac{1}{m^{1/2}}\right),
\end{equation}
where $\fNN$ is the fully converged NN trained with GD on the empirical squared loss and a sufficiently small learning rate. Inserting the result we obtained for $ \flin$ and setting $C = \max \{C_1,C_2\}$ finishes the proof.\QED

\begin{lemma}\label{lemma:reluexpression}
	For a fully-connected ReLU-NN $\fNN_{\theta}$ (\ref{NNparametrization}) with $L$ layers,  it holds for almost all $x \in  \mathbb{R}^d$ that
	\begin{equation}\label{eq:main-decomposition}
	\fNN_{\theta}(x)=\frac{1}{L}\langle \theta , \nabla_{\theta} \fNN_{\theta}(x)\rangle
	\end{equation}{}
\end{lemma}

\textbf{Proof.} Let $f^{NN}$ be a fully-connected NN as defined in the notation section. For simplicity we switch from the NTK parametrization to the \cite{he2015delving} initialization. The proof is equivalent in both parametrizations. The only difference is that we can suppress some of the scaling factors in the initialization of the weights.
The scalar product we use is the Hilbert-Schmid-product with $\langle A , B\rangle = tr\left[ A^TB\right]$  and $\theta = diag(W_{L},...,W_{1})$ is the diagonal matrix with all the weigh matrices on the diagonal. We restrict ourselves to networks with zero biases but the proof with non-zero biases is almost identical. 

To prove the lemma we can first proof that $\forall l \in \{1,...,L\}$ we have $f^{NN}_{W_l}(x) = \langle W_l , \nabla_{W_l} f(x)\rangle$. If this statement is true it is easy to see that Form.\ (\ref{eq:main-decomposition}) also holds true.

A ReLU-NN initialized with zero bias can be written in the following way
\begin{equation}
f^{NN}_W(x) = W_{L} \mathbf{l}_{h^{L-1}(x) \geq 0}W_{L-1} ... \mathbf{l}_{h^1(x)  \geq 0}W_{1}x,
\end{equation}
where $\mathbf{l}_{h^l(x) \geq 0}$ is the diagonal matrix with the step function on the diagonal elements corresponding to the components of $h^l(x) \geq 0$. Now, we can define $A(W_{L:l},h^l(x)) :=  W_{L} \mathbf{l}_{h^{L-1}(x) \geq 0}W_{L-1} ... \mathbf{l}_{ h^{l}(x) \geq 0}$ and $B(W_{l:1},x) :=  \mathbf{l}_{h^{l-1}(x) \geq 0}W_{l-1} ...$ to get
\begin{equation*}
f^{NN}_W(x) =A(W_{L:l},h^l(x)) W_l B(W_{l:1},x)
\end{equation*}
We can set $\frac{\partial (\mathbf{I}_{ h^l(x) = 0})_{ij}}{\partial (W_l)_{mn}} = 0$ as done in most of the literature and thus $\frac{\partial A_{ij}}{\partial (W_l)_{mn}} = 0$. Defining the derivative of the step function at the step to be zero is also done in most of the standard frameworks like tensorflow. Additionally, the region where the step function jumps is a $(d-1)$-dimensional subspace and thus the expression holds almost everywhere anyway. The derivative of $B(W_{l:1},x)$ with respect to $W_l$ is of course zero because it does not depend on $W_l$. 

Using this we can see that almost everywhere
\begin{equation*}
\langle W_l, \nabla_{W_l} f^{NN}(x)\rangle = \sum_{nm} (W_l)_{nm} \frac{\partial f^{NN}_W(x)}{\partial (W_l)_{nm}} = A(W_{L:l},x) W_l B(W_{l:1},x) =  f^{NN}(x),
\end{equation*}
Applying this result $L$-times, we get that almost everywhere
\begin{equation*}
f^{NN}_{\theta}(x) = \frac{1}{L}\langle \theta , \nabla_{\theta} \fNN_{\theta}(x)\rangle
\end{equation*}
\QED

\subsection{Proof of Thm.\ \ref{thm:initsigma} (lower bound)}\label{app:proof_initbelow}
\textbf{Proof.}
Let $\Phi(x,\sigma)$ be the feature vector of a NN (\ref{NNparametrization}) initialized with standard normal initialization with bias zero with variance $\sigma^2$. Then, it is easy to see that $\Phi(x,\sigma)  = \sigma^L \Phi(x,1)$ due to the homogeneity of the derivative w.r.t constants. Further we define
\begin{equation}
J(X_\tst,\sigma) :=  \frac{\| \Phi(X_\tst,\sigma)\left(\ii-\Phi(X,\sigma)^T(\Phi(X,\sigma)\Phi(X,\sigma)^T)^{-1} \Phi(X,\sigma)\right){\theta}_0 \|_2}{\sqrt{2N_{\tst}L}} 
\end{equation}
and $J(X_\tst) := J(X_\tst,1)$ and it is easy to see that $J(X_\tst)\geq 0$ almost surely. Further, due to the homogeneity of $\phi(x,\sigma)$ we get that $J(X_\tst,\sigma) = \sigma^LJ(X_\tst)$. On the other hand side 
\begin{align}
\phi(X_\tst,\sigma) \phi(X,\sigma)^T(\phi(X,\sigma)\phi(X,\sigma)^T)^{-1}Y &=	\phi(X_\tst,1)\phi(X,1)^T \nonumber \\ &\times (\phi(X,1)\phi(X,1)^T)^{-1}Y 
\end{align}
and therefore $\sqrt{L^{\normalfont\text{int}}_\tst} = \|f^{\normalfont\text{int}}(X_\tst)-Y\|/(\sqrt{2N_{\tst}})$ is independent of $\sigma$. Thus, applying the same arguments as in Thm.\ (\ref{thm:fNNfint}) to the whole test set we get with high probability over the random initialization that $\|\flin(X_\tst)-\fNN(X_\tst)\|_2 \leq O(1/\sqrt{m})$ and thus
\begin{align*}
\sqrt{L^\subsuperNN_\tst} &= \frac{1}{\sqrt{2N_{\tst}}}\|Y-\fNN(X_\tst)\|_2 \\
&\geq \frac{1}{2\sqrt{N_{\tst}}}\left[\|\fint(X_\tst)-\flin(X_\tst)\|_2- \|Y-\fint(X_\tst)\|_2 \right. \\ &- \left. \|\flin(X_\tst)-\fNN(X_\tst)\|_2 \right]\\
&\geq \sigma^{L} J(X_\tst) - \sqrt{L_\tst^{\normalfont\text{int}}}  -  O(1/\sqrt{m})
\end{align*}
\QED
\subsection{Proof of Thm.\ \ref{thm:initsigma} (upper bound)}\label{app:proof_boundinit}
\textbf{Proof.}
We start by applying the result of Thm.\ (\ref{thm:fNNfint}) to a the fixed test set. Thus, there exists $C_1= {\mathrm poly}(N,1/\delta,1/\lambda_0,N_\tst,1/\sigma) > 0 $ such that for all NNs with $m>C_1$ it holds with probability $1-\delta/3$ over the initialization that
\begin{align*}
\fNN(X_\tst)  &=  \frac{1}{L}\phi(X_\tst)\left[\ii-\phi(X)^T(\phi(X)\phi(X)^T)^{-1} \phi(X)\right]\theta_0\\
&+\phi(X_\tst) \phi(X)^T(\phi(X)\phi(X)^T)^{-1}Y +O\left(\frac{1}{m^{1/2}}\right)\cdot1,
\label{theo:keyformula}
\end{align*}
where $\phi(x)= \nabla_{\theta}\fNN_{\theta}(x)\big|_{\theta=\theta_0}$ and therefore $\frac{1}{\sqrt{2N_{\tst}}}\|\flin(X_\tst) - \fNN(X_\tst)\|_2 \leq O(1/\sqrt{m})$ using the expressions we have obtained for $\fNN$ and $\flin$ in the proof of Thm.\ \ref{thm:fNNfint} (see ep.\ ).
Next, we need to bound the difference between the linearized solution and the interpolating solution $\fint$ (App.\ \ref{app:interpolating_kernel}). The major difficulty here is to find a bound that is independent of $m$ or at least does not grow with $m$. This enables us to minimize the two errors simultaneously. The squared 2-norm of the difference between the linearized and the interpolating solution is given by
\begin{align*}
&\|\fint(X_\tst) - \flin(X_\tst)\|_2 = \|\frac{1}{L}\phi(X_\tst)\left[\ii-\phi(X)^T(\phi(X)\phi(X)^T)^{-1} \phi(X)\right]\theta_0\|_2 \\ 
\\ &\leq \|\frac{1}{L}\phi(X_\tst)\theta_0\|_2+\|\frac{1}{L}\phi(X_\tst)\phi(X)^T(\phi(X)\phi(X)^T)^{-1} \phi(X)\theta_0\|_2
\\&\leq \|\frac{1}{L}\phi(X_\tst)\theta_0\|_{2}+\|\phi(X_\tst)\phi(X)^T\|_\text{op}\|(\phi(X)\phi(X)^T)^{-1}\|_{\text{op}} \|\frac{1}{L}\phi(X)\theta_0\|_2
\end{align*}
Now, since $\phi(x)\phi(x^\prime)^T$ converges in probability to $\Theta(x,x^\prime)$ (defined in Sect.\ \ref{sec:notation}) there exists a $C_2={\mathrm poly}(N,1/\delta,1/\lambda_0,N_\tst,1/\sigma)$ such that for all $m>C_2$ with probability $1-\delta/3$ over the random initialization \cite{du2018gradient}
\begin{align*}
\|\Theta(X,X)- \phi(X)\phi(X)^T\|_{\text{op}} &\leq \lambda_0/2 \quad \text{and}\\
\|\Theta(X_\tst,X)- \phi(X_\tst)\phi(X)^T\|_{\text{op}} &\leq \sqrt{N_\tst}\lambda_0/2 
\end{align*}
Thus, we get
\begin{align*}
\|\phi(X_\tst)\phi(X)^T\|_\text{op}&\|(\phi(X)\phi(X)^T)^{-1}\|_{\text{op}} \|\frac{1}{L}\phi(X)\theta_0\|_2  \\ & \leq \left(\|\Theta(X_\tst,X)\|_\text{op}+\sqrt{N_\tst}\lambda_0/2\right)\frac{2}{\lambda_0} \|\frac{1}{L}\phi(X)\theta_0\|_2
\end{align*}
Next, we define $\lambda_{\text{sup}} := \|\Theta(X_\tst,X)\|_\text{op} $, which is independent of $m$, to arrive at
\begin{align*}
\|\phi(X_\tst)\phi(X)^T\|_\text{op}\|(\phi(X)\phi(X)^T)^{-1}\|_2 \|\frac{1}{L}\phi(X)\theta_0\|_2 &\leq \sqrt{N_\tst} \left(2\frac{\lambda_{\text{sup}}}{\lambda_0\sqrt{N_\tst}}+1\right) \\ &\times \|\frac{1}{L}\phi(X)\theta_0\|_2
\end{align*}
In the next step we can now make use of Lem.\ \ref{lemma:boundorthterm} to see that $\mathbb{E}\left[\|\frac{1}{L}\phi(X)\theta_0\|_2^2\right] \leq N\sigma^{2L}/(2^{L-1})$ and $\mathbb{E}\left[\|\frac{1}{L}\phi(X_\tst)\theta_0\|_2^2\right] \leq N_\tst\sigma^{2L}/(2^{L-1})$
and thus using Markov's inequality we get with probability $1-2\delta/3$ over the initialization that
\begin{equation*}
\frac{1}{2N_{\tst}}\|\fint(X_\tst) - \flin(X_\tst)\|_2^2 < \frac{ 3\sigma^{2L}}{(2^{L}\delta)}\left[1+  N\left(2\frac{\lambda_{\text{sup}}}{\lambda_0\sqrt{N_\tst}}+1\right)^2\right]
\end{equation*}

Now, setting $C = \max\{C_1,C_2\}$ and making use of the triangle inequality we get with probability $1-\delta$ over the initialization that
\begin{equation*}
\begin{aligned}
\sqrt{L^\subsuperNN_\tst} &= \frac{1}{\sqrt{2N_{\tst}}}\|Y-\fNN(X_\tst)\|_2 \\
&\leq \frac{1}{\sqrt{2N_{\tst}}}\left[ \|Y-\fint(X_\tst)\|_2+\|\fint(X_\tst)-\flin(X_\tst)\|_2 \right. \\  &+ \left.\|\flin(X_\tst)-\fNN(X_\tst)\|_2 \right] \\
&\leq \sqrt{L_\tst^{\normalfont\text{int}}} + O(\sigma^L)+O(1/\sqrt{m})
\end{aligned}
\end{equation*}
We believe that the bound we found for $\|\fint(X_\tst) - \flin(X_\tst)\|_2$ is rather conservative because we tried to find a bound which is independent of $m$. In practice this difference is usually small and closer to $\|\fNN_{\theta_0}\|_2$ and thus only becomes significant for large $\sigma$. 
\QED

\begin{lemma}\label{lemma:boundorthterm}
	For a fully-connected ReLU-NN $\fNN_\theta$ where we initialize all the weights randomly with $W_0^l\sim\mathcal{N}(0,\ii)$ and biases $b_0^l=0$ for all $l\in \{1,\cdots,L\}$,  it holds for all data matrices $X\in \mathbb{R}^{N\times d}$ with $N$ data points in $B_1^d(0)$, that $\mathbb{E}\left[ \|\fNN_{\theta_0}(X)\|_2^2)\right] \leq \frac{\sigma^{2L}}{2^{L-1}}N$.
\end{lemma}
\textbf{Proof.}
To start off we can again write the Neural Network as a product of matrices
\begin{equation*}
\fNN_{\theta_0}(x) = W_{L} \mathbf{l}_{h^{L-1}(x) \geq 0}W_{L} ... \mathbf{l}_{h^0(x)  \geq 0}W_{1}x.
\end{equation*}
Similar to the proof of Lem.\ \ref{lemma:reluexpression} we do not make the factors $\sigma/\sqrt{m}$ explicit but rather use the re-parametrization trick and initialize with $W^l_{ij}\sim\mathcal{N}(0,\sigma^2/m)$. This is purely out of convenience and the results are all unchanged.

We generalize an approach used by \cite{du2019width} to non-linear networks. This approach makes use of the fact that for a random matrix $A \in \mathbb{R}^{d_1\times d_2}$ with i.i.d. $\mathcal{N}(0,1)$ entries and an arbitrary non-zero vector $v \in \mathbb{R}^{d_2}-\{0\}$ we have $\mathbb{E}\left[\frac{\| Av \|^2}{\|v \|^2}\right] = d_1$, because $\frac{\| Av \|^2}{\|v \|^2}$  is $\chi_{d_1}^2$ distributed. Defining $\tilde{Z}_i =  \frac{\|\mathbf{l}_{h^{i}(x) >0} W_{i} \mathbf{l}_{h^{i-1}(x) \geq 0}W_{i-1} ... \mathbf{l}_{ h^1(x)  \geq 0}W_{1}x\|^2}{\|\mathbf{l}_{h^{i-1}(x) \geq 0}W_{i-1} ... \mathbf{l}_{h^1(x)  \geq 0}W_{1}x\|^2} $ and  $Z_i =  \frac{\|W_{i} \mathbf{l}_{h^{i-1}(x) \geq 0}W_{i-1} ... \mathbf{l}_{ h^1(x)  \geq 0}W_{1}x\|^2}{\|\mathbf{l}_{h^{i-1}(x) \geq 0}W_{i-1} ... \mathbf{l}_{h^1(x)  \geq 0}W_{1}x\|^2} $  one can rewrite:
\begin{equation*}
\mathbb{E}\left[ |\fNN_{\theta_0}(x)|^2\right] =  \mathbb{E}[Z_L\tilde{Z}_{L-1}\ldots \tilde{Z}_{1}]\|x\|_2^2 =\left(\prod_{l=1}^{L-1}\mathbb{E}[\tilde{Z}_l]\right)\mathbb{E}[Z_L]\|x\|_2^2,
\end{equation*}
where the expectations are taken over the initialized weights in each layer. In the last equality we use that the expectation over the weights of different layers become independent, because $\mathbb{E}\left[\frac{\| Av \|^2}{\|v \|^2}\right]$ with a random normally distributed matrix $A$ is independent of $v$. Thus, we can evaluate the expectations sequentially starting from the last layer to the first layer. Further, it is easy to see that $\mathbb{E}\left[\frac{\|\mathbf{l}_{Av  \geq 0} Av \|^2}{\|v \|^2}\right] = \frac{1}{2}\mathbb{E}\left[\frac{\|Av \|^2}{\|v \|^2}\right]$, which holds due to the symmetry of the normal distribution. Inserting this leads to
\begin{equation*}
\mathbb{E}\left[ |\fNN_{\theta_0}(x)|^2\right] = \frac{1}{2^{L-1}}\left(\prod_{l=1}^{L}\mathbb{E}[Z_l]\right)\|x\|_2^2 =  \frac{\sigma^{2L}}{2^{L-1}}\|x\|_2^2.
\end{equation*}
Applying this to the whole data matrix we get
\begin{equation*}
\mathbb{E}\left[ \|\fNN_{\theta_0}(X)\|_2^2\right] =  \frac{\sigma^{2L}}{2^{L-1}}\sum_{i=1}^{N}\|x_i\|_2^2 \leq \frac{\sigma^{2L}}{2^{L-1}}N,
\end{equation*}
where the last inequality follows because $\forall i \in \{1,\cdots,N\}$ we have $x_i \in B_1^d(0)$.

\begin{lemma}\label{lemma:boundorthterm_bias}
	For a fully-connected ReLU-NN $\fNN_\theta$ where we initialize all the weights and biases randomly with $\theta_0\sim\mathcal{N}(0,1)$,  it holds for all data matrices $X\in \mathbb{R}^{N\times d}$ with $N$ data point in $B_1^d(0)$, that $\mathbb{E}\left[ \|\fNN_{\theta_0}(X)\|_2^2)\right] \leq 2\sum_{i=1}^{L}\left(\frac{\sigma^2}{2}\right)^i N$.
\end{lemma}

\textbf{Proof.}
The proof is very similar to the proof of Lem.\ \ref{lemma:boundorthterm} but now we define $\tilde{Z}_i =  \frac{\|\mathbf{l}_{h^{i}(x) >0} h^{i}(x)\|^2_2}{\|h^{i-1}(x)\|_2^2} $ and  $Z_i =   \frac{\|h^{i}(x)\|^2_2}{\|h^{i-1}(x)\|_2^2} $ and again we get 
\begin{equation}\label{expation_formula}
\mathbb{E}\left[ |\fNN_{\theta_0}(x)|^2\right] =  \mathbb{E}[Z_L\tilde{Z}_{L-1}\ldots \tilde{Z}_{1}]\|x\|_2^2.
\end{equation}
Now, since the biases are not initialized to zero we can define $\tilde{h}^{i}(x):= \begin{pmatrix}h^{i}(x) \\ 1\end{pmatrix} \in \RR^{(m_i+1)} $ and $\tilde{W}_{i}(x):= \begin{pmatrix}W_{i}(x) , b^{i}\end{pmatrix}\in \RR^{m_i\times (m_{i-1}+1)}$ to get
\begin{align*}
\mathbb{E}\left[\tilde{Z_i}\right] &=  \mathbb{E}\left[\frac{\|\mathbf{l}_{h^{i}(x) >0} h^{i}(x)\|^2_2}{\|h^{i-1}(x)\|_2^2}\right] = \mathbb{E}\left[\frac{\|\mathbf{l}_{W_i h^{i-1}(x)+b >0} \left(W_i h^{i-1}(x)+b\right)\|^2_2}{\|h^{i-1}(x)\|_2^2}\right] \\
&= \mathbb{E}\left[\frac{\|\mathbf{l}_{\tilde{W}_i \tilde{h}^{i-1}(x) >0} \tilde{W}_i \tilde{h}^{i-1}(x)\|^2_2}{\|\tilde{h}^{i-1}(x)\|_2^2}\right]\frac{\|\tilde{h}^{i-1}(x)\|_2^2}{\|h^{i-1}(x)\|_2^2}\\
&=  \mathbb{E}\left[\frac{\|\mathbf{l}_{\tilde{W}_i \tilde{h}^{i-1}(x) >0} \tilde{W}_i \tilde{h}^{i-1}(x)\|^2_2}{\|\tilde{h}^{i-1}(x)\|_2^2}\right]\left(1+\frac{1}{\|h^{i-1}(x)\|_2^2}\right) \\ &\leq \frac{\sigma^2}{2}\left(1+\frac{1}{\|h^{i-1}(x)\|_2^2}\right),
\end{align*}
where in the last equality we used again the same insight as in Lem.\ \ref{lemma:boundorthterm} such that $\mathbb{E}\left[\frac{\|\tilde{W}_i \tilde{h}^{i-1}(x)\|^2_2}{\|\tilde{h}^{i-1}(x)\|_2^2}\right] \leq 1$. Plugging this in to eq.\ (\ref{expation_formula}) and again using that all the expectations become independent, we get
\begin{equation*}
\mathbb{E}\left[ |\fNN_{\theta_0}(x)|^2\right] = 2\sum_{i=1}^{L}\left(\frac{\sigma^2}{2}\right)^i\|x\|_2^2
\end{equation*}
and thus, again using that $\forall i \in \{1,\cdots,N\}$ we have $x_i \in B_1^d(0)$, we get
\begin{equation*}
\mathbb{E}\left[ \|\fNN_{\theta_0}(X)\|_2^2\right] =  2\sum_{i=1}^{L}\left(\frac{\sigma^2}{2}\right)^i\sum_{i=1}^{N}\|x_i\|_2^2 \leq 2\sum_{i=1}^{L}\left(\frac{\sigma^2}{2}\right)^iN,
\end{equation*}
\QED
\subsection{Proof of Thm.\ \ref{thm:generaladaptive}}\label{app:proof_generaladaptive}
\textbf{Proof.}
Let us again first consider the linear model and then in the end make use of Thm.\ (\ref{thm:adaptivelinear}) to show that this model is close to the full NN. Thus, we start by deriving the solution of $\flin_{\tilde{\theta}_t}$ which is the linearization where the weights are trained on the squared loss of the linearized model. Later, we will show that this solution is at most $O\left(\frac{1}{m^{1/2}}\right)$ apart from the full NN trained on the squared loss. We train the linear model, starting from the same weights $\theta_0$ as the full NN, w.r.t. the squared loss using the adaptive update step (\ref{adaptive_update})
\begin{equation}\label{eq:adaptiveupdatestep}
\tilde{\theta}_{t+1} = \left(\ii -\frac{\eta}{N} D_t \phi(X) \phi(X)^T\right)\tilde{\theta}_t+\frac{\eta}{N} D_t \phi(X)^T\hat{Y},
\end{equation}
where $\hat{Y} := Y-\fNN_{\theta_0}(X)+\phi(X)\theta_0$. We prove the following via induction
\begin{equation}\label{induction_hypothesis}
\tilde{\theta}_t = \theta_0 +A_t\left[ \ii - \prod_{z=t-1}^{0}(\ii - \frac{\eta}{N}\phi(X) D_z\phi(X)^T)\right]\left(Y-\fNN_{\theta_0}(X)\right)+B_t
\end{equation}
IA: For $t=0$ we can explicitly evaluate the adaptive update setp (\ref{eq:adaptiveupdatestep}) to arrive at $\tilde{\theta}_1 = \theta_0+\left(\eta/N\right)D_0 \phi(X)^T\left(Y-\fNN_{\theta_0}(X)\right)$ and by evaluating our induction hypothesis for $t=1$ we arrive at
\begin{equation*}
\tilde{\theta}_1 = \theta_0 +A_1\frac{\eta}{N}\phi(X)D_0\phi(X)^T\left(Y-\fNN_{\theta_0}(X)\right) = \theta_0+\frac{\eta}{N} D_0 \phi(X)^T\left(Y-\fNN_{\theta_0}(X)\right)
\end{equation*}

IS: We again make use of the adaptive update rule in equ.\ (\ref{eq:adaptiveupdatestep}) and the following properties
\begin{equation*}
\forall t \in  \mathbb{N}; \quad \phi(X)A_t = \ii \quad \text{and} \quad \phi(X)B_t = 0,
\end{equation*}
which are easy to verify, to arrive at
\begin{align*}
\tilde{\theta}_{t+1} &=  \left(\ii -\eta D_t \phi(X) \phi(X)^T\right)\tilde{\theta}_t+\frac{\eta}{N} D_t \phi(X)^T\hat{Y} \\
&=\left( \ii-\frac{\eta}{N} D_t\phi(X)^T\phi(X)\right) A_t\left[\ii - \prod_{i=t-1}^{0}(\ii - \frac{\eta}{N} \phi(X)D_i\phi(X)^T)\right]\\ &\times (Y-\fNN_{\theta_0}(X)) \\
&+ \left(\ii -\frac{\eta}{N} D_t \phi(X) \phi(X)^T\right)B_t+  \left(\ii -\frac{\eta}{N} D_t \phi(X) \phi(X)^T\right)\theta_0+\frac{\eta}{N} D_t \phi(X)^T\hat{Y}\\
&=\theta_0 + \left( A_t-\frac{\eta}{N} D_t\phi(X)^T\right) \left[\ii - \prod_{i=t-1}^{0}(\ii - \frac{\eta}{N} \phi(X)D_i\phi(X)^T)\right]\\ &\times (Y-\fNN_{\theta_0}(X))+ B_t + \frac{\eta}{N} D_t \phi(X)\left(Y-\fNN_{\theta_0}(X)\right)\\
&= \theta_0 + A_{t+1}\left( \ii-\frac{\eta}{N} \phi(X)D_t\phi(X)^T\right) \left[\ii - \prod_{i=t-1}^{0}(\ii - \frac{\eta}{N} \phi(X)D_i\phi(X)^T)\right]\\ &\times(Y-\fNN_{\theta_0}(X))+\frac{\eta}{N} D_t \phi(X)\left(Y-\fNN_{\theta_0}(X)\right)\\
&+(A_t-A_{t+1})\left[\ii -\prod_{i=t-1}^{0}\left(\ii-\frac{\eta}{N} \phi(X)D_i\phi(X)^T\right)\right](Y- \fNN_{\theta_0}(X))+B_t \\
&= \theta_0 +A_{t+1}\left[\ii -\prod_{i=t}^{0}\left(\ii-\frac{\eta}{N} \phi(X)D_i\phi(X)^T\right)\right](Y- \fNN_{\theta_0}(X)) +B_{t+1}
\end{align*}
This proves  our induction hypothesis (\ref{induction_hypothesis}). Plugging this result into $\flin_{\theta_{t}}(x)$ gives us
\begin{align*}
\flin_{\theta_{t}}(x) &=  \phi(x)A_{t}\left[\ii -\prod_{i=t}^{0}\left(\ii-\frac{\eta}{N} \phi(X)D_i\phi(X)^T\right)\right](Y- \fNN_{\theta_0}(X)) \\ &+\fNN_{\theta_0}(x)+\phi(x)B_{t}
\end{align*}
Next, similar to the non-adaptive regression we now make use of Thm.\ \ref{thm:adaptivelinear}, which states that if the adaptive matrices concentrate around some constant matrix $D\in \mathbb{R}^{P\times P}$ there exists $C = {\mathrm poly}(N,1/\delta,1/\lambda_0,1/\sigma)$ such that for all NN  where the number of hidden units $m$ is larger than $C$ we get $\forall x \in B^d_1(0)$ with probability $1-\delta$ over the initializations that
\begin{equation}
\sup_{t}|\fNN_{\theta_t}(x) - \flin_{\tilde{\theta}}(x)| \leq O\left(\frac{1}{m^{1/2}}\right),
\end{equation}
where $\fNN_{\tilde{\theta}_t}$ represents the NN where the weights $\tilde{\theta}$ are initialized also to $\theta_0$ and trained with adaptive GD. Inserting the result we obtained for $\flin$ therefore finishes the proof.
\QED
\subsection{Proof of Thm.\ \ref{thm:gd_sgd}}\label{app:proof_generalsgd}
We prove here the special case of $D_t= \ii$ but the more general case, where $D_t=D=const$, can be proven by exactly the same line of arguments by just substituting $\ii$ by $D$. 
Let $I = \{1,\cdots,N\}$ be the set of all the indices that label our data points in $X$. Further, we define  $\phi(X_C) := \mathcal{P}_C \phi(X)$, where $\mathcal{P}_C$ is a projection operator with 
\begin{equation*}
(\mathcal{P}_C)_{ij} = \delta_{ij} \ii_{i\in C} \quad \text{with} \quad \ii_{i\in C} := \begin{cases}
1 &\, \text{if} \quad i\in C \\
0 & \, \text{else}
\end{cases},
\end{equation*}
for some $C\subseteq I$. This notation also implies that $\phi(X_I)=\mathcal{P}_C\phi(X)+\left(\ii -\mathcal{P}_C\right)\phi(X)=\phi(X)$. Further, we use $B_i\subset I$ as the set of indices of all data points in the $i$-th mini-batch. Using the notation introduces above we can write the SGD update step as an adaptive method with 
\begin{equation*}
D_t:= \frac{N}{|B_t|}\left(\mathcal{P}_{B_t}\phi(X)\right)^T\left(\phi(X)\phi(X)^T\right)^{-1}\phi(X),
\end{equation*} 
because of
\begin{align*}
\theta_{t+1} &= \left(\ii - \frac{\eta}{N} D_t \phi(X)^T\phi(X) \right)\theta_t +\eta D_t \phi(X)^T\hat{Y}\\
&= \left(\ii - \frac{\eta}{|B_t|} \phi(X_{B_t})^T\phi(X) \right)\theta_t +\eta \phi(X_{B_t})^T\hat{Y}\\
&= \left(\ii - \frac{\eta}{|B_t|} \phi(X_{B_t})^T\phi(X_{B_t}) \right)\theta_t +\eta \phi(X_{B_t})^T\hat{Y}_{B_t}\\
&= \theta_t-\eta \nabla_{\theta}L_{\mathcal{D}_{B_t}}(\theta)|_{\theta= \theta_t}, 
\end{align*}
where $\hat{Y}:= Y-\fNN_{\theta_0}(X)+\phi(X)\theta_0$. The second last line follows because $\phi(X_{B_t})^T = (\mathcal{P}_{B_t}\phi(X_{B_t}))^T = \phi(X_{B_t})^T\mathcal{P}_{B_t}$ and the last line is exactly the SGD update step with $D_t= \ii$ defined in (\ref{generalsgdstep}).
This means that SGD can be analysed using the tools from the adaptive theorem (see Thm.\ \ref{thm:generaladaptive}). One dificulty of using Thm.\ \ref{thm:generaladaptive} is that the $A_t$'s are not well defined since the inverse of $(\phi(X)D_t\phi(X)^T)$ does not exist for our choice of $D_t$. Thus, to make full use of these results we first define
\begin{equation*}
D_t^\epsilon := \frac{N}{|B_t|}\left(\mathcal{P}_{B_t}^\epsilon\phi(X)\right)^T\left(\phi(X)\phi(X)^T\right)^{-1}\phi(X),
\end{equation*}
where
\begin{equation*}
\mathcal{P}_{B_t}^\epsilon := \mathcal{P}_{B_t}  + \epsilon\left(\ii-\mathcal{P}_{B_t}\right),
\end{equation*}
which is invertible for all $\epsilon >0$. Now, since $\mathcal{P}_{B_t}^\epsilon$ is invertible also the inverse of $(\phi(X)D_t^\epsilon \phi(X)^T)$ exists with $(\phi(X)D_t^\epsilon \phi(X)^T)^{-1} = \left(\mathcal{P}_{B_t}^\epsilon\right)^{-1}(\phi(X)\phi(X)^T)^{-1}$. From the corresponding update step
\begin{equation*}
\theta^\epsilon_{t+1} = \left(\ii - \frac{\eta}{N}\eta D_t^\epsilon \phi(X)^T\phi(X) \right)\theta_t^\epsilon +\frac{\eta}{N} D_t^\epsilon\phi(X)^T\hat{Y},
\end{equation*}
where again $\hat{Y}:= Y-\fNN_{\theta_0}(X)+\phi(X)\theta_0$, we can of course recover $\theta_t$ with $\theta_t = \lim_{\epsilon \to 0}\theta_t^\epsilon$. This limit is well defined since the update step has no inverce dependency on $\epsilon$.
Using these $D_t^\epsilon$ and $\mathcal{P}_{B_t}^\epsilon$ we can now analyse the $A^\epsilon_t$'s since now everything is well defined.
\begin{align*}
A^\epsilon_{t+1} &= D^\epsilon_t\phi(X)^T\left(\phi(X)D^\epsilon_t\phi(X)^T\right)^{-1}\\
&= \left(\mathcal{P}_{B_t}^\epsilon\phi(X)\right)^T\left(\phi(X)\left(\mathcal{P}_{B_t}^\epsilon\phi(X)\right)^T\right)^{-1}=  \phi(X)^T \mathcal{P}_{B_t}^\epsilon \left(\phi(X)\phi(X)^T\mathcal{P}_{B_t}^\epsilon\right)^{-1}\\
&=\phi(X)^T \mathcal{P}_{B_t}^\epsilon \left(\mathcal{P}_{B_t}^\epsilon\right)^{-1} \left(\phi(X)\phi(X)^T\right)^{-1} = \phi(X)^T\left(\phi(X)\phi(X)^T\right)^{-1}
\end{align*}
Thus, we can see that $A_{t+1}^\epsilon$ is completely independent of $\epsilon$ and the mini-batch $B_t$ we choose. Making use of the results we recovered in the proof of the general adaptive theorem for the linear model (see. Proof \ref{app:proof_generaladaptive} eq.\ \ref{induction_hypothesis}), we can see that $B_t = 0$ because the $A_t^\epsilon$'s are constant. Thus, evaluating the expression for $\epsilon \to 0$ we get
\begin{align*}
f^{lin_{SGD}}_{\theta_t}(x)=& \fNN_{\theta_0}(x)+ \phi(x)\phi(X)^T(\phi(X)\phi(X)^T)^{-1} \\
&\times\left[ \ii - \prod_{z=t-1}^{0}(\ii - \eta\phi(X) D_z\phi(X)^T)\right] (Y-\fNN_{\theta_0}(X))
\end{align*}

Now, to show that for sufficiently small learning rate $\eta$ our linear model converges to zero training error is more complicated than for GD. Therefore, we first focus on a single epoch, e.g. the first epoch, consisting of $E$ batches, all with the same batch size $b:= |B_k|$. Combining all data points of the $E$ batches gives us back the whole dataset or $\cup_{z=1}^{E}B_z = I$. First, we want to show that $\|\prod_{z=E}^{1}(\ii - \eta\phi(X) D_z\phi(X)^T)\|_{op} < 1-\frac{\eta}{2b} \lambda_{min}(\phi(X)\phi(X)^T)$ for sufficiently small $\eta$. Further, for $ S \subseteq \{1,\cdots,E\}$ we get
\begin{align*}
\|\prod_{z=E}^{1}(\ii - &\eta\phi(X) D_z\phi(X)^T)\|_{op} = \|\prod_{z=E}^{1}(\ii - \frac{\eta}{b}\phi(X)\phi(X)^T\mathcal{P}_{B_z})\|_{op} \\
&\leq \| \ii -\frac{\eta}{b} \phi(X)\phi(X)^T \sum_{z=1}^{E}\mathcal{P}_{B_z}\|_{\text{op}} +\sum_{z=2}^{E} \sum_{|S|=z} \prod_{i\in S}\frac{\eta}{b} \| \phi(X)\phi(X)^T\mathcal{P}_{B_i}\|_{\text{op}} \\
&\leq \| \ii -\frac{\eta}{b} \phi(X)\phi(X)^T\|_{\text{op}} +\sum_{z=2}^{E} \sum_{|S|=z} \prod_{i\in S}\frac{\eta}{b} \|\phi(X)\phi(X)^T\|_{\text{op}} \|\mathcal{P}_{B_i}\|_{\text{op}} \\
&= \| \ii -\frac{\eta}{b} \phi(X)\phi(X)^T\|_{\text{op}} +\sum_{z=2}^{E} \sum_{|S|=z} \prod_{i\in S}\frac{\eta}{b} \| \phi(X)\phi(X)^T\|_{\text{op}} \\
&\leq \| \ii -\frac{\eta}{b} \phi(X)\phi(X)^T\|_{\text{op}} +\sum_{z=2}^{E} \binom{E}{z} \left(\frac{\eta}{b} \| \phi(X)\phi(X)^T\|_{\text{op}}\right)^z,
\end{align*}
where in the second inequality we made use of $\sum_{z=1}^{E}\mathcal{P}_{B_z}= \ii$, since the mini-batches do not overlap, and for the third inequality we used $\|\mathcal{P}_{B_i}\|_\text{op} =1$. Next, similar than for GD, we can make use of the fact that there exits $C_1$ such that for all NNs with $m>C_1$ we have with high probability over the initialization that $\phi(X)\phi(X)^T$ is positive definite and thus for $\eta < \frac{b}{\lambda_{max}(\phi(X)\phi(X)^T)}$ we get
\begin{align*}
\|\prod_{z=E}^{1}(\ii - \eta\phi(X) D_z\phi(X)^T)\|_{op} &\leq \ii -\frac{\eta}{b} \lambda_{min}(\phi(X)\phi(X)^T) \\ &+\sum_{z=2}^{E} \binom{E}{z} \left(\frac{\eta}{b} \| \phi(X)\phi(X)^T\|_{\text{op}}\right)^z
\end{align*}

It is easy to see that the second term is negative and $O(\eta)$ and the third term is positive and $O(\eta^2)$. Therefore, by choosing a sufficiently small learning rate $\eta$, we get  $\|\prod_{z=i+E}^{i}(\ii - \eta\phi(X) D_z\phi(X)^T)\|_{op} < 1-\frac{\eta}{2b} \lambda_{min}(\phi(X)\phi(X)^T)<1$.
Further, we can apply this result to any epoch and thus see that $\prod_{z=t-1}^{0}(\ii - \eta\phi(X) D_z\phi(X)^T)$ is a product of $K$ matrices with operator norm smaller than $1-\frac{\eta}{b} \lambda_{min}(\phi(X)\phi(X)^T)$, where $K$ represents the number of epochs trained. Thus, for $K \to \infty$ we get $\prod_{z=t-1}^{0}(\ii - \eta\phi(X) D_z\phi(X)^T) \to 0$ and therefore
\begin{equation*}
f^{lin}_{SGD}(x)= \fNN_{\theta_0}(x)+ \phi(x)\phi(X)^T(\phi(X)\phi(X)^T)^{-1}(Y-\fNN_{\theta_0}(X))
\end{equation*}
Now, when training a NN with SGD, it is easy to see that one can recover a similar result as Lem.\ \ref{lemma:lee2019wide} using the results of Thm.\ 5 and 2 of \cite{allen2018convergence} and combining them with the argument used in Thm.\ H.1 of \cite{lee2017deep}. Using this result we see that there exists a $C_2= {\mathrm poly}(N,1/\delta,1/\lambda_0,1/\sigma) $ such that $\forall m \geq C_2$ and $\forall x \in B_1^d(0)$ it holds with probability $1-\delta$ over the initializations that
\begin{equation*}
f^{NN}_{SGD}(x)= \fNN_{\theta_0}(x)+ \phi(x)\phi(X)^T(\phi(X)\phi(X)^T)^{-1} (Y-\fNN_{\theta_0}(X))+O\left(\frac{1}{m^{1/2}}\right)
\end{equation*}
Thus, inserting the result of Thm.\ \ref{thm:fNNfint}, choosing $C=\max\{C_1,C_2\}$ and using the fact that both models were initialized to the same $\theta_0$ we get with probability $1-\delta$ over the initialization that
\begin{equation*}
|f^{NN}_{SGD}(x)-f^{NN}_{GD}(x)|= O\left(\frac{1}{m^{1/2}}\right)
\end{equation*}
\QED
\section{Linearization for adaptively trained NNs}\label{app:linearizationadaptive}
In this section we want to show with a basic constraint on the adaptive matrices, how one can extend the proof of \cite{lee2017deep} to NNs trained with adaptive optimization methods. This generalization is needed for the proof of Thm.\ \ref{thm:generaladaptive}. The idea is to make use of a concentration property of the adaptive matrices $D_t$. We say that a sequence of adaptive matrices concentrates around $D\in\RR^{P\times P}$ if there exists $Z \in \mathbb{R}$ such that $\| D_t-D\|_{\text{op}}/ D_{\max} \leq Z/\sqrt{m}$ holds for all $t\in\mathbb{N}$, where $D_{\max} := \sup_t \|D_t\|_{\text{op}}$; this is the simplest assumption under which we can generalize Thm.\ \ref{thm:fNNfint}. Making use of this property we can prove the following theorem
\begin{theorem}\label{thm:adaptivelinear}
	Denote the linearization of a NN (\ref{NNparametrization}) around its initialization $\theta_0$ by $\flin_\theta(x):=\fNN_{\theta_0}(x)+(\nabla_\theta\fNN_\theta(x)\big|_{\theta_0})(\theta-\theta_0)$. Further, for $t\geq0$, let $\tilde{\theta_t}$ and $\theta_t$ be the parameters obtained by adaptive gradient descent training with adaptivity matrix $D_t \in \mathbb{R}^{P\times P}$ starting from $\theta_0$ with sufficiently small step size $\eta$ of the full NN and its linearization respectively. Let the adaptivity matrix concentrate around some constant matrix $D\in \mathbb{R}^{P\times P}$. Then, there exists some $C = {\mathrm poly}(1/\delta,N,\lambda_0,1/\sigma)$ such that for all NNs with $m \geq C$ it holds with probability at least $1-\delta$ over the initialization that: $sup_{t} |\fNN_{\tilde{\theta_t}}(x) - \flin_{\theta_t}(x)|^2 \leq O(1/m)$ for all $x\in B_1^d(0)$.
\end{theorem}
In the proof of this theorem we are not concerned about the specific dependency of $C$ on $N$, $\lambda_0$, $\delta$ and $\sigma$ but rather want to show the basic idea behind the linearization and how one can generalize this idea to NNs trained with adaptive optimization methods. The condition we impose on the adaptivity matrices $D_t$ in our theorem are quite restrictive and we believe that one can not only loosen this restriction using different techniques but also show the advantages of adaptive optimization methods in convergence which are usually observed in practice. Since we are not concerned about the training error convergence we leave this open for future work.

Before we start we present Lem.\ 2 of \cite{lee2019wide}, which proves two key properties of NNs. These properties are independent of the training method and thus stay valid. We changed the notation of the lemma slightly to match the notation of this work.
\begin{lemma}\cite{lee2019wide}\label{lem:jacobian}
	For every $C >0$ there exists a $K >0$ such that for all NNs with $m\geq 2\ln(\frac{2}{\delta})$ it holds with probability $1-\delta$ over the random initialization that
	\begin{align}
	\| J(\theta)-J(\tilde{\theta})\|_F &\leq K \frac{\| \theta - \tilde{\theta}\|_2}{\sqrt{m}} \label{jacobian_formula} \\
	\| J(\theta)\|_F &\leq K,
	\end{align}
	for all $\theta, \tilde{\theta} \in B^M_{\theta_0}(Cm^{-1/2})$ and with $J(\theta) = \phi_{\theta}(X)$.\footnote{We added an additional factor of $\frac{1}{\sqrt{m}}$ to formula \ref{jacobian_formula}, which was  omitted from the original Lemma but looking at Lem.\ 1 of the same work one can easily see that this factor was missing.}
\end{lemma}

Next, we want to investigate the training dynamics of a NN trained with adaptive gradient descent with adaptive matrices $D_t$ instead of plain gradient descent. We show the following properties

\begin{theorem} \label{thm:trainerroradaptive}
	Let $\fNN_{\theta_t}$ be a NN as defined in (\ref{NNparametrization}), were the weights are updated according to the adaptive update step (\ref{adaptive_update}), with adaptive matrices $D_t\in \mathbb{R}^{P\times P}$ at step $t$, starting from $\theta_0$ with sufficiently small learning rate $\eta$. Let the adaptivity matrix concentrate around some constant matrix $D\in \mathbb{R}^{P\times P}$. Then, there exits $R_0 > 0$, $C = {\mathrm poly}(1/\delta,N,1/\lambda_0,1/\sigma)$ and $K>1$, such that for every $m \geq C$, with probability at least $(1-\delta)$ over the random initialization it holds:
	\begin{align}
	&\|g(\theta_t)\|_2 \leq \left( 1- \frac{\eta_0 \lambda_0}{3} \right)^t R_0  \label{firsteq}\\
	&\sum^t_{j=1}\|\theta_j -\theta_{j-1}\|_2  \leq K\eta_0 D_{max}\sum_{j=1}^{t}\left(1-\frac{\eta_0 \lambda_0}{3}\right)^{j-1}R_0 \leq \frac{3KR_0D_{max}}{\lambda_0}	\label{secondeq} \\
	& \sup_t\| \hat{\Theta}_0 - \hat{\Theta}_t\|_F \leq 2K^2\|D\|_{op}\left(\frac{3KR_0 D_{max}}{\lambda_0}+Z\right) m^{-1/2}  \label{thirdeq},	
	\end{align}
	where $\hat{\Theta}_t = \phi_{\theta_t}(X)D_t\phi_{\theta_t}(X)^T$, $\Theta = \lim_{m \to \infty} \hat{\Theta}$, with $\hat{\Theta}= \phi_{\theta_0}(X)D\phi_{\theta_0}(X)^T$ and $\lambda_0 = \lambda_{min}(\Theta)$, $\lambda_{max} = \lambda_{max}(\Theta)$ and $g(\theta_t) = \fNN_{\theta_t}(X)-Y$.
\end{theorem} 
\textbf{Proof.}
We start by making use of Lem.\ \ref{lemma:boundorthterm} where we showed that $ \|\fNN_{\theta_t}(X)\|_2^2 \leq \frac{N\sigma^{2L}}{\delta 2^{L-1}}$ with probability larger than $1-\delta$ over the initialization. Using this, it is easy to see that with probability $1-\delta/3$ there exists a $R_0(1/\delta,N,\sigma,L)$ such that
\begin{equation*}
\|g(\theta_0)\|_2 \leq R_0 \quad , \forall m \in \mathbb{N}
\end{equation*}
Next, we start by proving equations \ref{firsteq} and \ref{secondeq} and set $C = \frac{3KR_0}{\lambda_0}$. We prove these relations by induction. The induction start is trivial, thus we proceed by assuming the statement holds for some arbitrary $t\in \mathbb{N}$ and then by induction we get
\begin{equation*}
\|\theta_{t+1} -\theta_t\|_2 \leq \eta \|D_t\|_{op}\|J(\theta_t)\|_{op}\| g(\theta_t)\|_2 \leq K\eta_0 D_{max}\left(1-\frac{\eta_0 \lambda_0}{3}\right)^tR_0,
\end{equation*}
which also implies
\begin{equation*}
\|\theta_{t+1}-\theta_0\|_2 \leq \frac{3KR_0D_{max}}{\lambda_0}.
\end{equation*}
Now, we can proceed to eq.\ \ref{firsteq}. We have
\begin{align*}
\|g(\theta_{t+1})\|_2 &= \|g(\theta_{t+1})-g(\theta_{t})+g(\theta_{t})\|_2 \\
&= \|J(\tilde{\theta}_t)\left(\theta_{t+1}-\theta_t\right)  +g(\theta_{t})\|_2\\
&= \|-\eta J(\tilde{\theta}_t)D_tJ(\theta_t)g(\theta_{t})+g(\theta_{t})\|_2\\
&\leq \| 1 -\eta J(\tilde{\theta}_t) D_t J(\theta_t)\|_{op}	\left( 1- \frac{\eta_0 \lambda_0}{3} \right)^t R_0,
\end{align*}
where in the second equality we have made use of the component wise mean value theorem. It now remains to show that with probability $1-\delta/2$ over the initialization
\begin{equation*}
\| 1 -\eta J(\tilde{\theta}_t) D_t J(\theta_t)\|_{op} \leq \left(1- \frac{\eta \lambda_0}{3} \right)
\end{equation*}
We know that $\hat{\Theta} \to \Theta$ in probability and thus there exists $m_1 \in \mathbb{N}$ such that with probability $1-\delta/2$ over the initialization we have $\forall m >m_1$ that
\begin{equation*}
\|\hat{\Theta} - \Theta\|_F \leq \frac{\eta \lambda_0}{3}.
\end{equation*}
Therefore, we get
\begin{align*}
\| 1 -\eta J(\tilde{\theta}_t) D_t J(\theta_t)\|_{op} &\leq  \|1 - \eta \Theta\|_{op}+ \eta\|\Theta-\hat{\Theta}\|_{op} \\ &+\eta\|J(\theta_0)DJ(\theta_0)^T-J(\tilde{\theta_t})D_tJ(\theta_t)^T\|_{op}\\
&\leq 1-\eta\frac{2}{3}\lambda_0 +\eta\|J(\theta_0)DJ(\theta_0)^T-J(\tilde{\theta_t})D_tJ(\theta_t)^T\|_{op},
\end{align*}
where in the last line we use that for $\eta < \frac{2}{\lambda_0+\lambda_{max}}$ we get $\|1-\eta \Theta\|_{op} \leq 1-\eta \lambda_0$. Further, making use of Lem.\ \ref{lem:jacobian} and the concentration constraint of the $D_t$ matrices we get
\begin{align*}
&\eta\|J(\theta_0)DJ(\theta_0)^T-J(\tilde{\theta_t})D_tJ(\theta_t)^T\|_{op} \\
&\leq \|J(\theta_0)D\left(J(\theta_0)^T-J(\theta_t)^T\right) \\ &+\left(J(\theta_0)-J(\tilde{\theta_t})\right)D_tJ(\theta_t)^T +J(\theta_0)(D-D_t)J(\theta_t)\|_{op} \\
&\leq \eta K^2 D_{max} \left[\left(1+\frac{\|D\|_{op}}{D_{max}}\right)\|\theta_t-\theta_{0}\|_2 +\frac{\|D-D_t\|_{op}}{D_{max}}\right]\\
&\leq \eta K^2 D_{max} \left[\left(1+\frac{\|D\|_{op}}{D_{max}}\right)\frac{3KR_0}{\lambda_0 \sqrt{m}}+\frac{Z}{\sqrt{m}}\right].
\end{align*}
Thus, if we choose 
\begin{equation*}
m \geq m_2 := \left[\left(1+\frac{\|D\|}{D_{max}}\right)\frac{3KR_0}{\lambda_0}+Z\right]^2\frac{9K^4D_{max}^2}{\lambda_0^2}
\end{equation*}
with probability $1-\delta/2$ over the initialization we have $ \| 1 -\eta J(\tilde{\theta}_t) D_t J(\theta_t)\|_{op} \leq \frac{\eta \lambda_0}{3}$.
Now, combining these result by setting $C \geq \max\{m_1,m_2\}$, it holds with probability larger than $1-\delta$ over the random initialization that
\begin{equation*}
\|g(\theta_{t+1})\|_2 \leq \left( 1- \frac{\eta \lambda_0}{3} \right)^{t+1} R_0
\end{equation*}
Now, that we have shown the first two equations we can now move to eq.\ \ref{thirdeq} which follows using the same arguments
\begin{align*}
&\| \hat{\Theta}_0 - \hat{\Theta}_t\|_F = \| J(\theta_0)D J(\theta_0)^T-J(\theta_t)D_tJ(\theta_t)^T + J(\theta_0)\left(D-D_0\right) J(\theta_0)^T\\ &+ J(\theta_t)\left(D-D_t\right) J(\theta_t)^T\|_F\\
&\leq \|J(\theta_0)\|_{op}\|D\|_{op} \|J(\theta_0)-J(\theta_t)\|_{F}+ \|J(\theta_0)-J(\theta_t)\|_{op}\|D\|_{op}\|J(\theta_0)\|_{F} \\
&+ \|J(\theta_0)\|_{op}\|\left(D-D_0\right)\|_{op} \|J(\theta_0)^T\|_F+ \|J(\theta_t)\|_{op}\|\left(D-D_t\right)\|_{op} \| J(\theta_t)^T\|_F \\
&\leq 2K^2 \|D\|_{op} \frac{\left(\|\theta_0-\theta_t\|_2+Z\right)}{\sqrt{m}} \leq 2K^2\|D\|_{op}\left(\frac{3KR_0 D_{max}}{\lambda_0}+Z\right) m^{-1/2} 
\end{align*}
were in the last line we made use of eq.\ \ref{firsteq}. This closes the proof.
\QED

\textbf{Proof. of Thm.\ \ref{thm:adaptivelinear}}
Now, that we have proven the adaptive version of theorem G.4 of \cite{lee2019wide} we can prove the same statement as in Thm. H.1 of \cite{lee2019wide} for adaptivly trained NNs, substituting the standard GD kernels with the new kernels $\hat{\Theta} = \phi_{\theta_t}(X)D\phi_{\theta_t}(X)^T$ and  $\Theta = \lim_{m \to \infty} \hat{\Theta}$. The proof of Thm. H.1 does not change due the changed kernels, since it solely relies on the properties shown in Thm.\ \ref{thm:trainerroradaptive} which we have shown for the NNs trained with adaptive optimization methods. Thus, the proof itself stays unchanged. This result then closes the proof of Thm. \ref{thm:adaptivelinear}. \QED

\end{document}